\documentclass[preprint,12pt]{elsarticle}
\usepackage{amssymb}
\graphicspath{{fig/}}
\usepackage{ifpdf}
\ifpdf
  \usepackage{epstopdf}
\fi
\usepackage[cmex10]{amsmath}
\usepackage{amsfonts}
\usepackage{bm}
\usepackage[colorlinks=true]{hyperref}
\usepackage[caption=false, font=footnotesize]{subfig}

\newcommand{\dataset}{{\cal D}}

\newcommand{\bigO}[0]{\mathcal{O}}
\newcommand{\GP}[0]{\mathcal{GP}}

\newcommand{\Normal}[0]{\mathcal{N}}
\newcommand{\vect}[1]{{\boldsymbol{\mathbf{#1}}}} % vector
\newcommand{\mat}[1]{{\boldsymbol{\mathbf{#1}}}} % matrix
\renewcommand{\d}[0]{\text{d}} % derivative
\newcommand{\prob}{p}

\newcommand{\LVB}[0]{\mathcal{L}_\text{VB}}
\newcommand{\LKL}[0]{\mathcal{L}_\text{CorrVB}}
\newcommand{\DKL}[0]{\text{KL}}
\newcommand{\cut}[1]{} % cut out a part of the text
\newcommand{\chol}[0]{\operatorname{chol}}

\journal{Pattern Recognition}

\begin{document}

\begin{frontmatter}

%% Title, authors and addresses

%% use the tnoteref command within \title for footnotes;
%% use the tnotetext command for the associated footnote;
%% use the fnref command within \author or \address for footnotes;
%% use the fntext command for the associated footnote;
%% use the corref command within \author for corresponding author footnotes;
%% use the cortext command for the associated footnote;
%% use the ead command for the email address,
%% and the form \ead[url] for the home page:
%%
%% \title{Title\tnoteref{label1}}
%% \tnotetext[label1]{}
%% \author{Name\corref{cor1}\fnref{label2}}
%% \ead{email address}
%% \ead[url]{home page}
%% \fntext[label2]{}
%% \cortext[cor1]{}
%% \address{Address\fnref{label3}}
%% \fntext[label3]{}

\title{Overlapping Mixtures of Gaussian Processes for the Data Association Problem}

%% use optional labels to link authors explicitly to addresses:
%% \author[label1,label2]{<author name>}
%% \address[label1]{<address>}
%% \address[label2]{<address>}

\author[unican]{Miguel L\'azaro-Gredilla\corref{cor1}}
\ead{miguellg@gtas.dicom.unican.es}
\author[unican]{Steven~Van~Vaerenbergh}
\ead{steven@gtas.dicom.unican.es}
\author[sheffield]{Neil Lawrence}
\ead{N.Lawrence@sheffield.ac.uk}
\cortext[cor1]{Corresponding author: Tel: +34 942200919 ext 802, Fax: +34 942201488.}

\address[unican]{Dept.\ Communications Engineering,
University of Cantabria,
39005 Santander, Spain }

\address[sheffield]{Dept.\ of Computer Science,
University of Sheffield,
S1 4DP Sheffield, UK}

\begin{abstract}
%% Text of abstract
In this work we introduce a mixture of GPs to address the data
  association problem, i.e.\ to label a group of observations according
  to the sources that generated them. Unlike several previously
  proposed GP mixtures, the novel mixture has the distinct
  characteristic of using no gating function to determine the
  association of samples and mixture components. Instead, all the GPs
  in the mixture are global and samples are clustered following
  ``trajectories'' across input space. We use a non-standard variational Bayesian
  algorithm to efficiently recover sample labels and learn the hyperparameters. We show how
  multi-object tracking problems can be disambiguated and also explore the
  characteristics of the model in traditional regression
  settings.
\end{abstract}

\begin{keyword}
%% keywords here, in the form: keyword \sep keyword
Gaussian Processes \sep Marginalized Variational Inference \sep Bayesian Models
%% MSC codes here, in the form: \MSC code \sep code
%% or \MSC[2008] code \sep code (2000 is the default)

\end{keyword}

\end{frontmatter}

%%
%% Start line numbering here if you want
%%
% \linenumbers

%% main text
\section{Introduction}

\begin{figure*}[!htb]
\centerline{\subfloat[One-dimensional observations.]{\includegraphics[width=6.1cm]{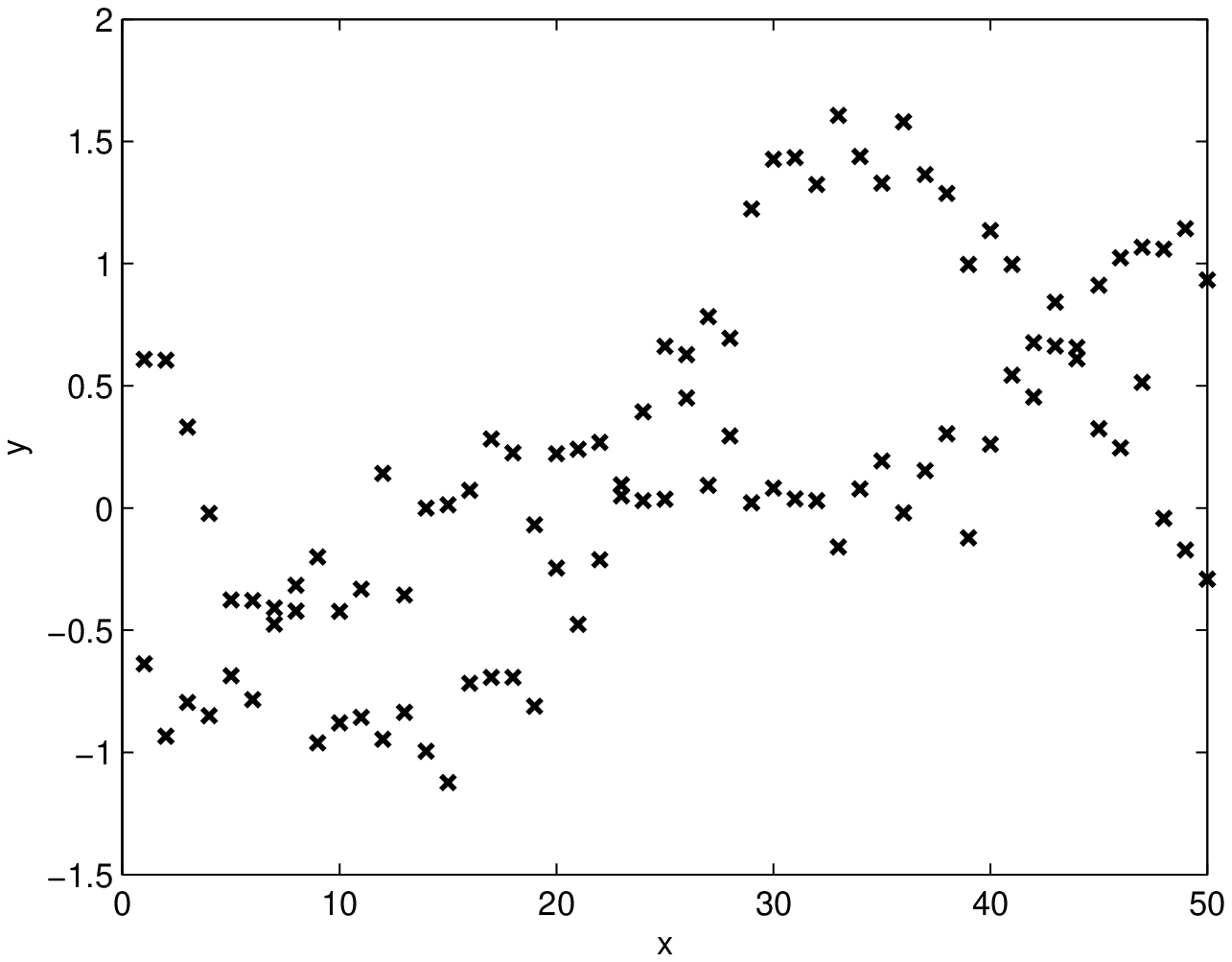}
}
\hfil
\subfloat[Solution obtained by the proposed method.]{\includegraphics[width=6.1cm]{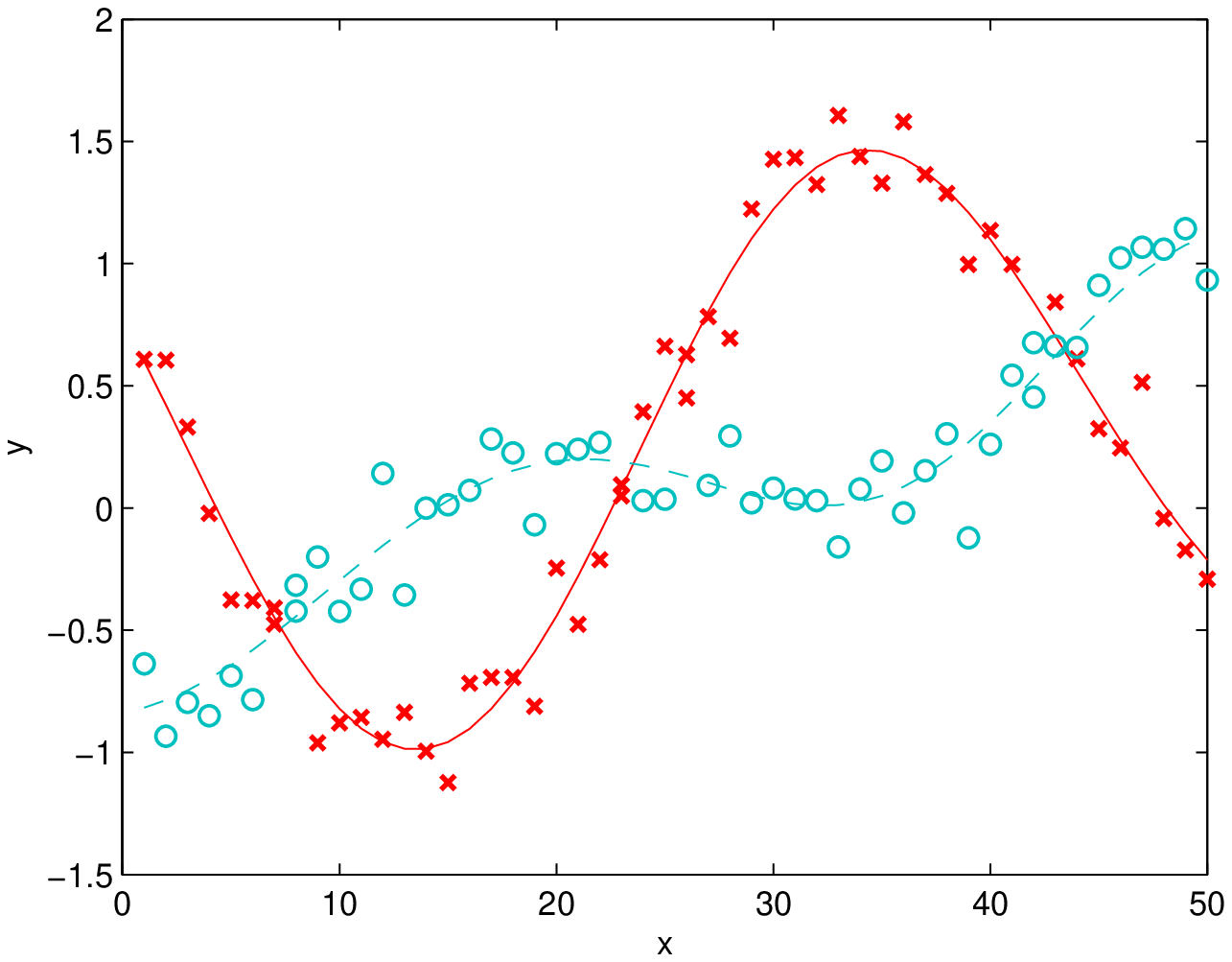}
}}
\caption{Example of a multi-target tracking scenario. Data association aims to identify what observations correspond to each source. %In this specific case, one observation of each source is available at each time instant, and the sampling interval is fixed.
}
\label{fig:candy}
\end{figure*}

The data association problem arises in multi-target tracking scenarios. Given a set of observations that represent the positions of a number of moving sources, such as cars or airplanes, data association consists of inferring which observations originate from the same source \cite{bar1987tracking,cox1993review}. Data association is found in tracking problems for instance in computer vision \cite{ullman1979interpretation}, surveillance, sensor networks \cite{singh2011} and radar tracking \cite{fortmann83jpdaf}. An example of data association with two sources is illustrated in Figure \ref{fig:candy}.

For a human observer, little effort is required to distinguish two noisy trajectories in this example, representing the paths followed by two objects in time. In this specific case, one observation of each target is available at each time instant, and the measurement instants are equally spaced in time, although neither of these properties are required in general.

Typical multi-target tracking algorithms operate online. They include joint Kalman filters \cite{Reid79multitarget} and joint particle filters \cite{karlsson2001montecarlo}. Given the predicted positions of the targets and a number of candidate observed positions, they usually make instant data association decisions based on nearest-neighbor criteria or statistically more sophisticated approaches such as the Joint Probabilistic Data-Association Filter (JPDAF) \cite{fortmann83jpdaf,karlsson2001montecarlo} or the Multiple Hypothesis Tracker (MHT) \cite{Reid79multitarget}. An important disadvantage of these classical techniques is that they usually require to determine a large number of parameters. This drawback motivated the development of several conceptually simpler approaches based on motion geometry heuristics \cite{cox1993review,chetverikov1999feature,veenman2001motion}. However, these approaches are usually limited to specific scenarios, and they show difficulties in the presence of noise and when several trajectories cross each other.

Most data association techniques can be significantly improved by postponing decisions until enough information is available to exclude ambiguities \cite{cox1993review}, although this causes the number of possible trajectories to grow exponentially. Some attempts have been made to restrain this combinatorial explosion, including the heuristic methods from  \cite{nagarajan87combinatorial,cox96efficient}. % sliding window approaches: see Veenman.

In this paper we present an algorithm based on Gaussian Processes that is able to consider all available data points in batch form whilst avoiding the exponential growth in potential tracks. As a result, it is capable to deal with difficult data association problems in which trajectories come very close and even cross each other. Furthermore, the algorithm does not require any knowledge about the model underlying the data, and it does not need time instants to be evenly spaced, nor to contain observations from all sources.

Gaussian Processes (GPs) \cite{bookce} are a powerful tool for Bayesian nonlinear regression. When combined in mixture models, GPs can be applied to describe data where there are local non-stationarities or discontinuities \cite{tresp, infmix, altinfmix, varmix}. The components of the mixture model are GPs and the prior probability of any given component is typically provided by a gating function. The role of the gating function is to dictate which GP is a priori most likely to be responsible for the data in any given region of the input space, i.e.,
the gating network forces each component of the GP mixture to be localized.

In this work we follow a different approach, inspired by the data association problem. In particular, for any given location in input space there may be multiple targets, perhaps corresponding to multiple objects in a tracking system. We are interested in constructing a GP mixture model that can associate each of these targets with separate components. When there is ambiguity, the posterior distribution of targets will reflect this. We therefore propose a simple mixture model in which each component is global in its scope. The assignment of the data to each GP is performed sample-wise, independently of input space localization. In other words, no gating function is used. We call this model the Overlapping Mixture of GPs (OMGP).

It has been brought to our attention that the proposed model bears resemblance with the work of \cite{malditos}. However, the focus of application is clearly different. In \cite{malditos}, the objective is to cluster a set of trajectories according to their similarity, whereas in this work we tackle the task of clustering observations into trajectories  (a more demanding task, since only single observations, as opposed to full trajectories, are available). Also, \cite{malditos} uses a standard variational Bayesian algorithm, whereas in this work we take advantage of non-standard variational algorithms \cite{klcorrected,miguelvhgp} to derive a tighter bound.

The remainder of this paper is organized as follows: In Section \ref{sec:review} we provide a brief review of GPs in the regression setting. Section \ref{sec:omgp} first introduces the OMGP model and then discusses how to perform efficient learning, hyperparameter selection, and predictions using this model. Experiments on several data sets are provided in Section \ref{sec:experiments}. We wrap up in Section \ref{sec:conclusions} with a brief discussion.

\section{Brief Review of Gaussian Processes}
\label{sec:review}

In recent years, Gaussian Processes (GPs) have attracted a lot of attention due to their nice analytical properties and their state-of-the-art performance in regression tasks (see \cite{thesisrasmussen}). In this section we provide a brief summary of the main results for GP regression, see \cite{bookce} for further details.

Assume that a set of $N$ multi-dimensional inputs and their
corresponding scalar outputs,
$\dataset\equiv\{\vect{x}_n,y_n\}_{i=1}^m$, are available. The
regression task is, given a new input $\vect{x}_*$, to obtain the
predictive distribution for the corresponding observation $y_*$ based
on $\dataset$.

The GP regression model assumes that the observations can be modeled as some noiseless latent function of the inputs plus independent noise
$ y\;=\;f(\vect{x})+\varepsilon$,
and then sets a zero-mean\footnote{To make this assumption hold, the sample mean of the set $\{y(\vect{x}_n)\}_{n=1}^m$ is usually subtracted from data before proceeding further.} GP prior on the latent function $f(\vect{x})\;\sim\;\GP(0, k(\vect{x},\vect{x}'))$ and a Gaussian prior on $\varepsilon\;\sim\;\Normal(0,\;\sigma^2)$ on the noise, where $k(\vect{x},\vect{x}')$ is a covariance function and $\sigma^2$ is a hyperparameter that specifies the noise power.

The covariance function $k(\vect{x},\vect{x}')$ specifies the degree of coupling between $y(\vect{x})$ and $y(\vect{x}')$, and it encodes the properties of the GP such as power level, smoothness, etc. One of the best-known covariance functions is the anisotropic squared exponential. It has the form of an unnormalized Gaussian,
$k(\vect{x},\vect{x}')\;=\;\sigma_0^2\exp\left(-\frac{1}{2}\vect{x}^\top\mat{\Lambda}^{-1}\vect{x}\right)$
and depends on the signal power $\sigma_0^2$ and the length-scales $\mat{\Lambda}$, where $\mat{\Lambda}$ is a diagonal matrix containing one length-scale per input dimension. Each length-scale controls how fast the correlation between outputs decays as the separation along the corresponding input dimension grows.
We will collectively refer to all kernel parameters as $\vect{\theta}$.

The joint distribution of the available observations (collected in $\vect{y}$) and some unknown output $y(\vect{x}_*)$ is a multivariate Gaussian distribution, with parameters specified by the covariance function:
\begin{equation}
 \left[\!\!
  \begin{array}{c}
   \vect{y} \\
   y_*
  \end{array}
  \!\!\right]
 \;\sim\;
 \Normal\left( \vect{0},\;\left[\!\!
   \begin{array}{cc}
    \mat{K}+\sigma^2\mat{I}_N & \vect{k}_*\\
    \vect{k}_*^\top & k_{**}+\sigma^2\\
   \end{array}
   \!\!\right]\right)\;,
   \label{eq:jointprior}
\end{equation}
where $[\mat{K}]_{nn'}=k(\vect{x}_n,\vect{x}_{n'})$, $[\vect{k}_{*}]_n=k(\vect{x}_n,\vect{x}_*)$ and $k_{**}=k(\vect{x}_*,\vect{x}_*)$. $\mat I_N$ is used to denote the identity matrix of size $N$. The notation $[\mat{A}]_{nn'}$ refers to entry at row $n$, column $n'$ of $\mat{A}$. Likewise, $[\vect{a}]_n$ is used to reference the $n$-th element of vector $\vect{a}$.

From \eqref{eq:jointprior} and conditioning on the observed
training outputs we can obtain the predictive distribution
\begin{align}
 \label{eq:preddist} \prob_{\text{GP}}(y_*|\vect{x}_*,\dataset)=\Normal(y_*|\mu_{\text{GP}*},\sigma_{\text{GP}*}^2)\\
 \mu_{\text{GP}*} = \vect{k}_{*}^\top (\mat{K}+\sigma^2\mat{I}_N)^{-1}\vect{
y} ~~~~
\nonumber \sigma_{\text{GP}*}^2 &= \sigma^2+k_{**}-
   \vect{k}_{*}^\top (\mat{K}+\sigma^2\mat{I}_N)^{-1}\vect{k}_{*}\;,
\end{align}
which is computable in $\bigO(N^3)$ time, due to the inversion\footnote{Of course, in a practical implementation, this inversion should never be performed explicitly, but through the use of the Cholesky factorization and the solution of the corresponding linear systems, see \cite{bookce}.} of the $N\times N$ matrix $\mat{K}+\sigma^2\mat{I}_N$.

Hyperparameters $\{\vect{\theta},\sigma\}$ are typically selected by maximizing the marginal likelihood (also called ``evidence'') of the observations, which is
\begin{equation}
\label{eq:logevidence}
 \log p(\vect{y}|\vect{\theta},\sigma) =
-\frac{1}{2} \vect{y}^\top \left(\mat{K} + \sigma^2\mat{I}_{N}\right)^{-1} \vect{y}
-\frac{1}{2}|\mat{K} + \sigma^2\mat{I}_N|
 -\frac{N}{2}\log(2\pi)\;.
\end{equation}

If analytical derivatives of \eqref{eq:logevidence} are available, optimization can be carried out using gradient methods, with each gradient computation taking $\bigO(N^3)$ time. GP algorithms can typically handle a few thousand data points on a desktop PC.

When dealing with multi-output functions, instead of a single set of
observations $\vect{y}$, $D$ sets are available, $\vect{y}_1\ldots
\vect{y}_D$, each corresponding to a different output dimension. In
this case we can assume independence across the outputs and perform
the above procedure independently for each dimension.
This will provide reasonable results for most problems, but if correlation between different dimensions is expected, we can take advantage of this knowledge and model them jointly using multi-task covariance functions \cite{bonilla}.

\section{Overlapping Mixtures of Gaussian Processes (OMGP)}
\label{sec:omgp}

% In this section we will introduce the Overlapping Mixture of Gaussian Processes (OMGP) model and show how to efficiently perform ference with it.

The overlapping mixture of Gaussian processes (OMGP) model assumes that there exist $M$ different latent functions $\{f^{(m)}(\vect{x})\}_{m=1}^M$ (which we will call ``trajectories''), and that each output is produced by evaluating one of these functions at the corresponding input and by adding Gaussian noise to it. The association between samples and latent functions is determined by the $N\times M$ binary indicator matrix $\mat{Z}$: Entry $[\mat{Z}]_{nm}$ being non-zero specifies that $n$-th data point was generated using trajectory $m$. Only one non-zero entry per row is allowed in $\mat{Z}$.

To model multi-dimensional trajectories (i.e., when the mixture model has multiple outputs), $D$ latent functions per trajectory can be used $\{f^{(m)}_d(\vect{x})\}_{m=1,d=1}^{M,D}$. Note that there is no need to extend $\mat{Z}$ to specifically handle the multi-output case, since all the outputs corresponding to a single input are the same data point and must belong to the same trajectory.

For convenience we will collect all the outputs in a single matrix $\mat{Y}=[\vect{y}_1\ldots\vect{y}_D]$ and all the latent functions of trajectory $m$ in a single matrix $\mat{F}^{(m)}=[\vect{f}_1^{(m)}\ldots\vect{f}_D^{(m)}]$. We will refer to all the latent functions as $\{\mat{F}^{(m)}\}$.

Given the above description, the likelihood of the OMGP model is
\begin{equation}
 \label{eq:likelihood}
p(\mat{Y}|\{\mat{F}^{(m)}\}, \mat{Z}) = \prod_{n=1,m=1,d=1}^{N,M,D} \Normal([\mat{Y}]_{nd}| [\mat{F}^{(m)}]_{nd},\sigma^2)^{[\mat{Z}]_{nm}}\;.
\end{equation}

Following the standard Bayesian framework, we place priors on the unobserved latent variables
\begin{equation}
\label{eq:prior}
p(\mat{Z}) = \prod_{n=1,m=1}^{N,M} [\mat{\Pi}]_{nm}^{[\mat{Z}]_{nm}}, ~~~~~~~~~
p(\mat{F}^{(m)}|\mat{X}) = \prod_{m=1,d=1}^{M,D} \Normal(\vect{f}^{(m)}_d|\vect{0},\mat{K}^{(m)})\;,
\end{equation}
i.e., a multinomial distribution over the indicators (in which $\sum_{m=1}^{M} [\mat{\Pi}]_{nm}=1~~\forall_n$) and independent GP priors over each latent function.\footnote{If correlation between different trajectories is known to exist, trajectories can be jointly modeled as a single GP, using a covariance function that accounts for this dependence. This would increase the computational complexity of inference for this model, but the following derivations can still be applied.} We allow different covariance matrices for each trajectory. Though the multinomial distribution is specified here in its more general form, additional constraints are usually imposed, such as holding the prior probabilities constant for all data points. For the sake of clarity, we will omit the conditioning on the hyperparameters $\{\vect{\theta}, \mat{\Pi}, \sigma^2\}$, which can be assumed to be known for the moment.

Unfortunately, the analytical computation of the posterior distribution $p(\mat{Z},\{\mat{F}^{(m)}\}|\mat{X}, \mat{Y})$ is intractable, so we will resort to approximate techniques.

\subsection{Variational approximation}
\label{sec:varapprox}

If the hyperparameters are known, it is possible to approximately compute the posterior using a variational approximation. We can use Jensen's inequality to construct a lower bound on the marginal likelihood as follows:
\begin{equation}
\label{eq:evidence}
\log p(\mat{Y}|\mat{X}) = \log \int p(\mat{Y}|\{\mat{F}^{(m)}\}, \mat{Z})p(\mat{Z}) \prod_{m=1}^M p(\mat{F}^{(m)}|\mat{X}) \d\{\mat{F}^{(m)}\} \d\mat{Z}
\end{equation}
$$
\geq \int q(\{\mat{F}^{(m)}\}, \mat{Z}) \log \frac{p(\mat{Y}|\{\mat{F}^{(m)}\}, \mat{Z})p(\mat{Z}) \prod_{m=1}^M p(\mat{F}^{(m)})|\mat{X})}{q(\{\mat{F}^{(m)}\}, \mat{Z})} \d\{\mat{F}^{(m)}\} \d\mat{Z}
=\LVB.
$$

Here $\LVB$ is a lower bound on $\log p(\mat{Y}|\mat{X})$ for any \emph{variational distribution} $q(\{\mat{F}^{(m)}\}, \mat{Z})$ and equality is attained if and only if $q(\{\mat{F}^{(m)}\}, \mat{Z}) = p(\mat{Z},\{\mat{F}^{(m)}\}|\mat{X}, \mat{Y})$. Our objective is therefore to find a variational distribution that maximizes $\LVB$, and thus becomes an approximation to the true posterior.
We will restrict our search to variational distributions that factorize as $q(\{\mat{F}^{(m)}\}, \mat{Z}) = q(\{\mat{F}^{(m)}\})q(\mat{Z})$.

If we assume that $q(\{\mat{F}^{(m)}\})$ is given (and therefore, also the marginals $q(\vect{f}^{(m)}_d) = \Normal(\vect{f}^{(m)}_d|\vect{\mu}^{(m)}_d,\mat{\Sigma}^{(m)})$ are available), it is possible to analytically maximize $\LVB $ with respect to $q(\mat{Z})$ by setting its derivative to zero and constraining it to be a probability density. The optimal $q(\mat{Z})$ is then:
\begin{equation}
\label{eq:estep}
q(\mat{Z}) = \prod_{n=1,m=1}^{N,M} [\hat{\mat{\Pi}}]_{nm}^{[\mat{Z}]_{nm}}\text{ with } [\hat{\mat{\Pi}}]_{nm} \propto
[{\mat{\Pi}}]_{nm}\exp(a_{nm})
\end{equation}
$$
\text{ with }~~~~
%\left\langle \log p([\mat{y}_d]_{n}|[\mat{f}^{(m)}_d]_{n}, [\mat{Z}]_{nm}) \right\rangle_{q([\mat{f}^{(m)}_d]_{n})} =
a_{nm} = \sum_{d=1}^D \left(
-\frac{1}{2\sigma^2}\left( ([\mat{y}_d]_{n}-[\mat{\mu}^{(m)}_d]_{n})^2+[\mat{\Sigma}^{(m)}]_{nn}\right)-\frac{1}{2}\log(2\pi\sigma^2)
\right),
 %\int q(\{\mat{F}^{(m)}\}, \mat{Z}) \log p(\mat{Y}|\{\mat{F}^{(m)}\}, \mat{Z}, \sigma^2) \d\{\mat{F}^{(m)}\} \d\mat{Z}
$$
where we see that the (approximate) posterior distribution over the indicators $q(\mat{Z})$ factorizes for each sample.

Analogously, assuming $q(\mat{Z})$ as known, it is possible to analytically obtain the distribution over the latent functions that maximizes $\LVB$. For the OMGP model, this distribution factorizes both over trajectories and dimensions, and is given by
\begin{subequations}
\label{eq:mstep}
\begin{equation}
q(\vect{f}^{(m)}_d) = \Normal(\vect{f}^{(m)}_d|\vect{\mu}^{(m)}_d,\mat{\Sigma}^{(m)})
\end{equation}
\begin{equation}
\text{ with }\mat{\Sigma}^{(m)}=(\mat{K}^{-1(m)}+\mat{B}^{(m)})^{-1}
\text{ and }\vect{\mu}^{(m)}_d=\mat{\Sigma}^{(m)} \mat{B}^{(m)} \vect{y}^{(m)}_d
\end{equation}
\end{subequations}
where $\mat{B}^{(m)}$ is a diagonal matrix with elements $[\hat{\mat{\Pi}}]_{1m}/\sigma^2\ldots [\hat{\mat{\Pi}}]_{Nm}/\sigma^2$.

It is now possible to initialize $q(\mat{Z})$ and $q(\vect{f}^{(m)}_d)$ from their prior distributions and iterate updates \eqref{eq:estep}  and \eqref{eq:mstep} to obtain increasingly refined approximations to the posterior. Since both steps are optimal with respect to the distribution that they compute, they are guaranteed to increase $\LVB$, and therefore the algorithm is guaranteed to converge to a local maximum.

Monotonous convergence can be monitored by computing $\LVB$ after each update. $\LVB$ can be expressed as
\begin{align*}
\LVB & = \left\langle \log p(\mat{Y}|\{\mat{F}^{(m)}\}, \mat{Z}) \right\rangle_{q(\{\mat{F}^{(m)}\}, \mat{Z})} \\
& - \DKL(q(\{\mat{F}^{(m)}\})||p(\{\mat{F}^{(m)}\}))-\DKL(q(\mat{Z})||p(\mat{Z}))
\end{align*}
where the first term is given by
$$
\left\langle \log p(\mat{Y}|\{\mat{F}^{(m)}\}, \mat{Z}) \right\rangle_{q(\{\mat{F}^{(m)}\}, \mat{Z})} =
\sum_{n,m}^{N,M}[\hat{\mat{\Pi}}]_{nm}a_{nm}\;,
$$
and the two remaining terms are the Kullback-Leibler (KL) divergences from the approximate posterior to the prior, which are straightforward to compute.

Update \eqref{eq:estep}  takes only $\bigO(NM)$ computation time, whereas \eqref{eq:mstep}  takes $\bigO(MN^3)$ time, due to the $M$ matrix inversions. The presented model therefore has the same limitations as conventional GPs regarding the size of the data sets that it can be applied to. However, when the posterior probability of some indicator $[\hat{\mat{\Pi}}]_{nm}$ is close to zero, sample $n$ no longer affects trajectory $m$ and can be dropped in its computation, thus reducing the cost. Furthermore, it is possible to use sparse GPs\footnote{Such as the standard FITC approximation, described in \cite{snelson06sparse} or the variational approach introduced in \cite{titsias}.} to reduce this cost\footnote{Obviously, the cost also depends on the quality of the approximation by a constant factor. If the FITC approximation with $r$ pseudo-inputs (or other rank-$r$ approximation) is used, the computational complexity could be expressed as $\bigO(MNr^2)$.} to $\bigO(MN)$ time by making use of the matrix inversion lemma.

\subsection{An improved variational bound for OMGP}
\label{sec:OMGPBound}

So far we have assumed that all the hyperparameters of the model are known. However, in practice, some procedure to select them is needed. The most straightforward way of achieving this would be to select them so as to maximize $\LVB$, interleaving this procedure with updates \eqref{eq:estep}  and \eqref{eq:mstep}. However, when the quality of this bound is sensitive to changes of the model hyperparameters, this approach results in very slow convergence. A solution to this problem is described in \cite{klcorrected} where the advantages of maximizing an alternative, tighter bound on the likelihood are shown.

The improved bound proposed in \cite{klcorrected} is still a lower bound on the likelihood but it can be proved that it is also an upper bound on the standard variational bound $\LVB$. As shown in \cite{klcorrected}, if we subtract $\LVB$ from the improved bound, the result takes on the form of a KL-divergence. This fact can be used both to show that it upper-bounds $\LVB$ (since KL-divergences are always positive) and to name the new bound, which is referred to as the KL-corrected variational bound.

The KL-corrected bound for the OMGP model arises when the term $\log \int p(\mat{Y}|\{\mat{F}^{(m)}\}, \mat{Z}) p(\mat{Z})\d\mat{Z}$ from the true marginal likelihood \eqref{eq:evidence} is replaced with $\int q(\mat{Z}) \log \frac { p(\mat{Y}|\{\mat{F}^{(m)}\}, \mat{Z}) p(\mat{Z})}{q(\mat{Z})}\d\mat{Z}$, which according to Jensen's inequality, constitutes a lower bound for any distribution $q(\mat{Z})$:
\begin{align*}
\log p(\mat{Y}|\mat{X}) & \geq \log \int \prod_{m=1}^M p(\mat{F}^{(m)}|\mat{X})
e^{ \int q(\mat{Z}) \log \frac { p(\mat{Y}|\{\mat{F}^{(m)}\}, \mat{Z}) p(\mat{Z})}{q(\mat{Z})}\d\mat{Z} }
 \d\{\mat{F}^{(m)}\} =\\
\LKL & = \sum_{m=1, d=1}^{M, D} \log \Normal(\vect{y}^{(m)}_d|\vect{0}, \mat{K}^{(m)}+\mat{B}^{-1(m)})\\
& - \DKL(q(\mat{Z})||p(\mat{Z}))
+ \frac{D}{2}\sum_{n=1, m=1}^{N, M} \log\frac{(2\pi\sigma^2)^{1-[\hat{\mat{\Pi}}]_{nm}}}{[\hat{\mat{\Pi}}]_{nm}}\;.
\end{align*}

The KL-corrected lower bound $\LKL$ can be computed analytically and has the advantage with respect to $\LVB$, of depending only on $q(\mat{Z})$ (and not $q(\{\mat{F}^{(m)}\})$), since it is possible to integrate $\prod_{m=1}^M p(\mat{F}^{(m)}|\mat{X})$ out analytically.

Bound $\LKL$ can be alternatively obtained by following the recent work in \cite{miguelvhgp} and optimally removing $q(\{\mat{F}^{(m)}\})$ from the standard bound. In the context of that work, $\LKL$ is referred to as the ``marginalized variational bound'', and it is made clear that $\LKL$ corresponds simply to $\LVB$ when, for a given  $q(\mat{Z})$, the optimal choice for $q(\{\mat{F}^{(m)}\})$ is made. In other words, for the same set of hyperparameters and the same $q(\mat{Z})$, if one choses $q(\{\mat{F}^{(m)}\})$ according to \eqref{eq:mstep}, both $\LVB$ and $\LKL$ would provide the same result.

Thus, learning is performed simply by optimizing $\LKL$ with respect to $q(\mat{Z})$ and the hyperparameters, iterating the following two steps:
\begin{itemize}
\item	E-Step: Updates \eqref{eq:estep} and \eqref{eq:mstep} are alternated, which monotonically increase both $\LVB$ and $\LKL$, until convergence. Hyperparameters are kept fixed.
\item	M-Step:	Gradient descent of $\LKL$ with respect to all hyperparameters is performed. Distribution $q(\mat{Z})$ is kept fixed.
\end{itemize}

Note that it is in the M-step where $\LKL$ becomes actually useful, since this improved bound remains more stable across different hyperparameter selections, due to it not depending on $q(\{\mat{F}^{(m)}\})$, as demonstrated in \cite{klcorrected}.

Of course, any strategy that maximizes $\LKL$ is valid, but we have found the above EM procedure to work well in practice.

%Unlike other clustering algorithms and mixture models, in the OMGP model increasing the number of trajectories (clusters) does not necessarily increase the marginal likelihood, since the parameters defining each trajectory have been integrated out, so $\LKL$ can also be used to select $M$.

Computing $\LKL$ according to the provided expression without incurring in numerical errors can be challenging in practice, since several inversions, which maybe unstable, are needed. Also, note that $\mat{B}^{(m)}$ can take arbitrarily small values and thus direct inversion may not be possible. An implementation-friendly expression where explicit inverses are avoided is
\begin{align*}
\LKL &= \sum_{m=1}^M
\Big(
-\frac{1}{2}\sum_{d=1}^D||\mat{R}^{(m)\top}\backslash(\mat{B}^{(m)\frac{1}{2}}\vect{y}^{(m)}_d)||^2
-D\sum_{n=1}^N \log [\mat{R}^{(m)}]_{nn}
\Big)\\
&- \DKL(q(\mat{Z})||p(\mat{Z}))
- \frac{D}{2}\sum_{n=1, m=1}^{N, M} [\hat{\mat{\Pi}}]_{nm}\log(2\pi\sigma^2)\;,
\end{align*}
where
$$
\mat{R}^{(m)} = \chol(\mat{I} + \mat{B}^{(m)\frac{1}{2}}\mat{K}^{(m)}\mat{B}^{(m)\frac{1}{2}})
$$
and the backslash has the usual meaning of solution to a linear system.\footnote{Expressions of the type $\mat{C}\backslash \vect{c}$ refer to the solution of the linear system $\mat{C}\vect{x} = \vect{c}$ and are a numerically stable operation requiring only $\bigO(N^2)$ time when $\mat{C}$ is triangular, which is the case here.}

\subsection{Predictive distributions}

The OMGP model can be used for a variety of tasks. In the data association problem (i.e., clustering data into trajectories) the task at hand is to cluster observations into trajectories, which can be achieved by assigning each observation to the trajectory that more likely generated it, i.e., to assign label $m^* = \arg\max_m [\hat{\mat{\Pi}}]_{nm}$ to the $n$-th observation, so no further computations are necessary. For other tasks, however, it can be necessary to obtain predictive distributions over the output space at new locations. Under the variational approximation, this predictive distributions can be computed analytically.

The predictive distribution in the output dimension $d$ corresponding to a new test input location $\vect{x}_*$ can be expressed as
\begin{align*}
p(y_{*d}| \vect{x}_*, \mat{X}, \mat{Y})
&=\sum_{m=1}^M [\mat{\Pi}]_{*m} \int p~(y_{*d}| \vect{f}^{(m)}_{d}, \vect{x}_*, \mat{X})~~p~(\vect{f}^{(m)}_{d}|\mat{X}, \mat{Y})
\d \vect{f}^{(m)}_{d}\\
&\approx
\sum_{m=1}^M [\mat{\Pi}]_{*m} \int p~(y_{*d}| \vect{f}^{(m)}_{d}, \vect{x}_*, \mat{X})~~ q~(\vect{f}^{(m)}_{d}|\mat{X}, \mat{Y})
\d \vect{f}^{(m)}_{d}\\
&= \sum_{m=1}^M [\mat{\Pi}]_{*m} \Normal(y_{*d} | \mu_{*d}^{(m)}, \sigma_{*d}^{2(m)})
\end{align*}
with
\begin{align*}
\mu_{*d}^{(m)} &= \vect{k}_*^{\top(m)} ~ (\mat{K}^{(m)}+\mat{B}^{(m)-1})^{-1}~\vect{y}_d,\\
 \sigma_{*d}^{2(m)} &= \sigma^2 + k_{**} - \vect{k}_*^{\top(m)}~ (\mat{K}^{(m)}+\mat{B}^{(m)-1})^{-1}~\vect{k}_*^{(m)} \;,
\end{align*}
i.e., a Gaussian mixture under the approximate posterior. The mixing factors $[\mat{\Pi}]_{*m}$ are the prior probabilities of each component, one of the given hyperparameters of the model, and typically constant for all inputs.

Note the correspondence of these predictive equations with the standard predictions for GP regression \eqref{eq:preddist}. The only difference is the noise component, which is scaled for each sample according to $[\hat{\mat{\Pi}}]_{nm}^{-1}$. In particular, as the posterior probability of a sample belonging to the current trajectory (sometimes known as ``responsibility'') decays, the amount of noise associated to that sample is proportionally grown, thus reducing its effect on the posterior process.

Due to the reasons mentioned in the previous subsection, the predictive equations should not be implemented directly. Instead, the following numerically-stable expressions should be used:
\begin{align*}
\mu_{*d}^{(m)} &= \vect{k}_*^{\top(m)} ~ \mat{B}^{(m)\frac{1}{2}}
(\mat{R}^{(m)}\backslash(\mat{R}^{(m)\top}\backslash(\mat{B}^{(m)\frac{1}{2}}\vect{y}^{(m)}_d))),\\
\sigma_{*d}^{2(m)} &= \sigma^2 + k_{**}
- ||\mat{R}^{(m)\top}\backslash(\mat{B}^{(m)\frac{1}{2}}\vect{k}_*^{\top(m)})||^2\;.
\end{align*}

\subsection{Batch versus online operation}

Though the description of OMGP is oriented towards batch data association tasks, this model can also be successfully applied to online tasks, by using a data set that grows over time. New samples are included as they arrive and the learning process is re-started, initializing it from the state that was obtained as a solution for the previous problem. Depending on the constraints of a given problem, many different optimizations can be made to avoid an explosion in computational effort, such as using low-rank updates.

Note, however, that since in this model all the elements in each latent function form a fully connected graph, the Markovian property does not hold and the computation time required for each update is not  constant. A possible workaround to achieve constant-time updates is to use constat-size data sets, for instance corresponding to a sliding window, and then perform low-rank updates to include and remove samples. However, we will not pursue that option in this work.

\section{Experiments}
\label{sec:experiments}

In this section we investigate the behavior of OMGP both in data association tasks and regression tasks, showing the versatility of this model. We use an implementation of OMGP in Matlab on a $3$GHz, dual-core desktop PC with $4$GB of memory, yielding executions times of the order of seconds for each experiment.

\subsection{Data association tasks}

\subsubsection{Toy data}

\begin{figure*}[ht]
\centerline{\subfloat[Observed rotating sources.]{\includegraphics[width=6.5cm]{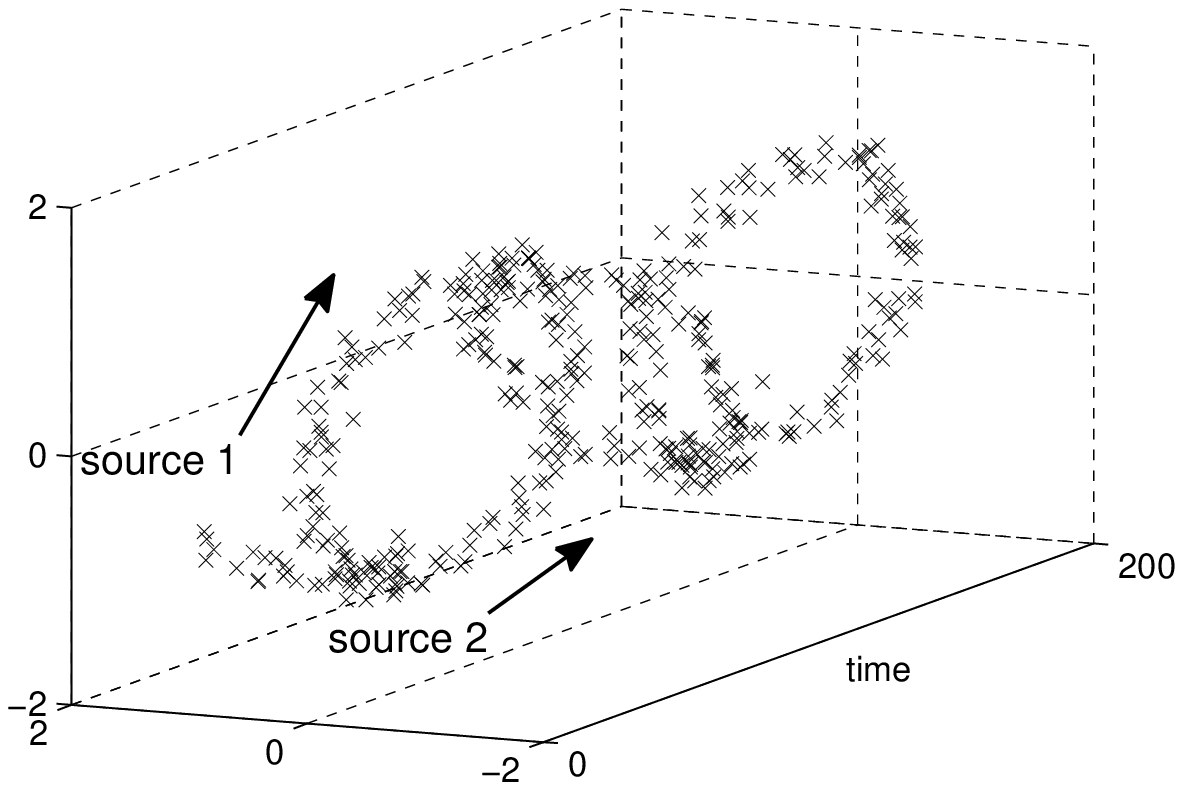} } \hfil \subfloat[Identified trajectories in
time.]{\includegraphics[width=6.5cm]{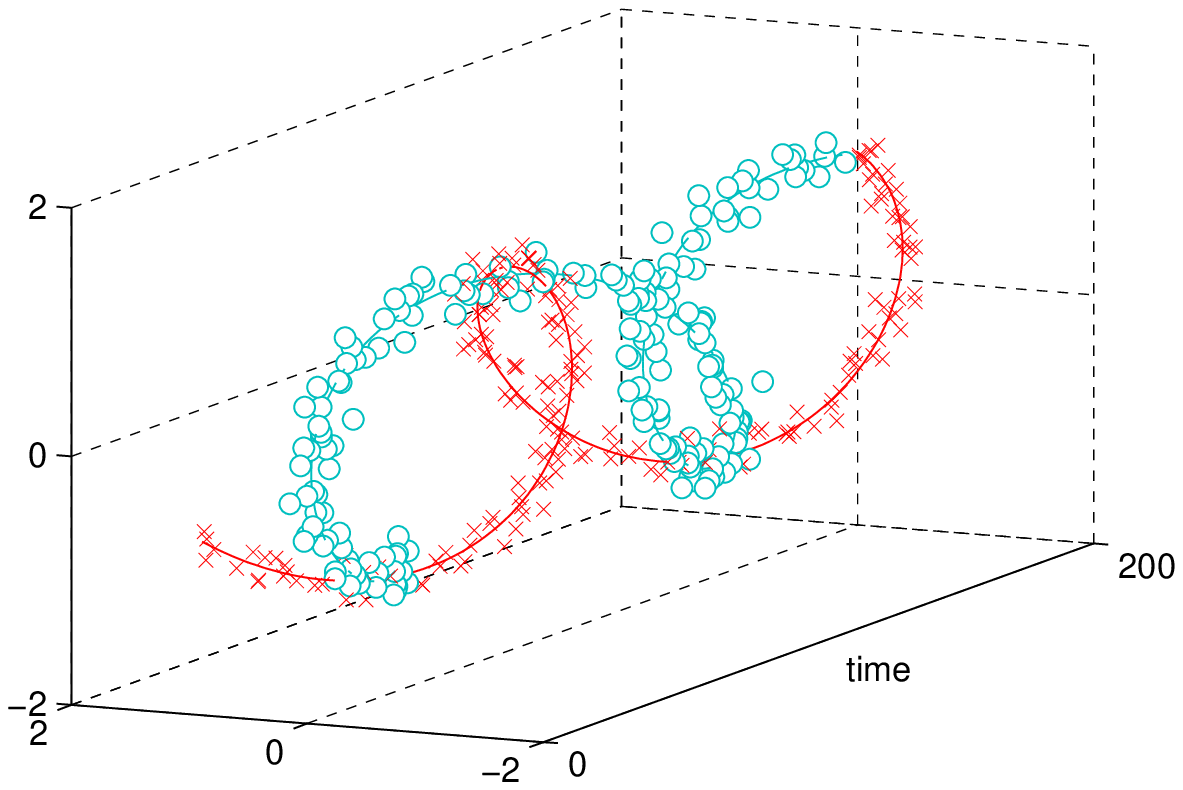}}
}
\centerline{\subfloat[Posterior probability of correct labels.]{\includegraphics[width=0.8\linewidth]{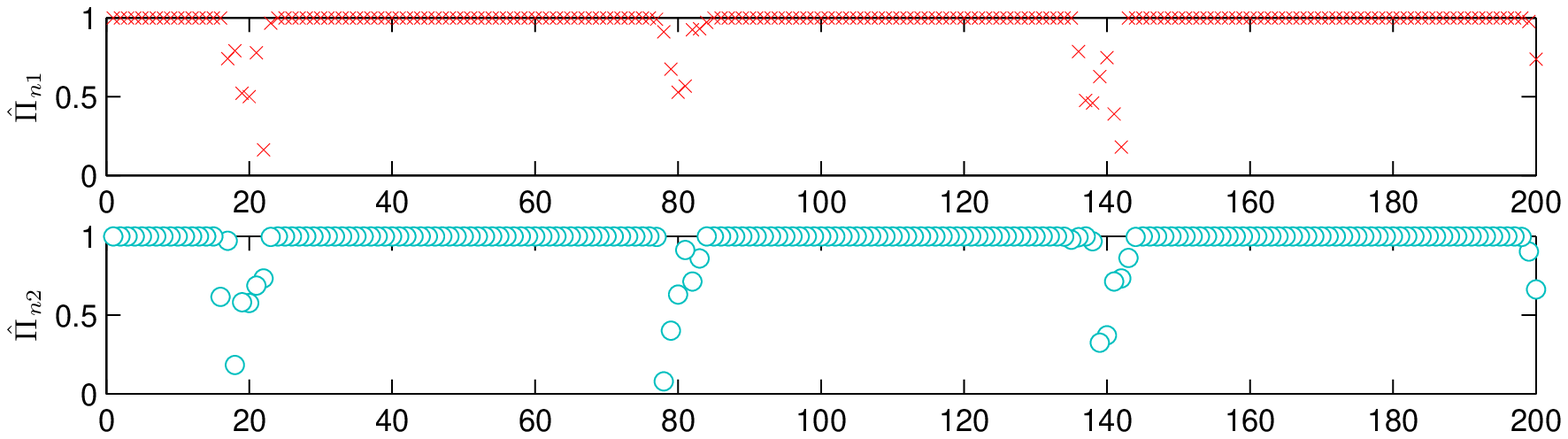}}}

\caption{(a) Observations for two sources that move in opposite circles. (b) The data association solution obtained by OMGP. (c) Posterior probability of the correct label for observations coming from source $1$ (top) and $2$ (bottom).}
\label{fig:springs}
\end{figure*}

We first apply OMGP to perform data association on a toy data set. The sources perform circular motions, one clockwise and one counterclockwise, as depicted in Fig.~\ref{fig:springs}(a). The available observations represent the measured positions of the sources (which include Gaussian noise) at known time instants. However, it is not known which observed position corresponds to which source. Since both trajectories are circles with the same center and radius, the sources cross each other twice per revolution, making the clustering problem more difficult. However, as shown in Fig.~\ref{fig:springs}(b), OMGP is capable of successfully identifying the unknown trajectories. Fig.~\ref{fig:springs}(c) illustrates the uncertainty about the estimated labels. Specifically, it shows a decrease in the posterior probability of the correct labels whenever the two sources come close.

\subsubsection{Missile-to-air multi-target tracking}

Next, we consider a missile-to-air tracking scenario as described in \cite{karlsson2001montecarlo}. The motion dynamics of this scenario are defined by the following state-space equations:
$$
\vect{s}_{t+1} = \begin{bmatrix}
\mat{I}_3 & T \mat{I}_3\\
\mat{O}_3 & \mat{I}_3\\
\end{bmatrix} \vect{s}_t + \begin{bmatrix}
\frac{T^2}{2} \mat{I}_3\\
T \mat{I}_3\\
\end{bmatrix} \vect{v}_t\\; \quad
\vect{r}_t = h(\vect{s}_t) = \begin{bmatrix}
\sqrt{X_t^2 + Y_t^2 + Z_t^2}\\
\arctan(\frac{Y_t}{X_t})\\
\arctan(\frac{-Z_t}{\sqrt{X_t^2+Y_t^2}})
\end{bmatrix} + \vect{e}_t.
$$
In this model, the state vector $\vect{s}_t = [X_t, Y_t, Z_t, V_{x,t}, V_{y,t}, V_{z,t}]$ contains the source position and velocity components, $\vect{r}_t$ contains the observed measurements, $T$ is the sampling interval, and $\mat{I}_3$ and $\mat{O}_3$ represent the $3\times 3$ unity matrix and null matrix, respectively. The process noise $\vect{v}_t$ and measurement noise $\vect{e}_t$ are assumed Gaussian, $\vect{v}_t \in \Normal(0,\mat{Q})$ and $\vect{e}_t \in \Normal(0,\mat{R})$. For more details refer to \cite{karlsson2001montecarlo}. The problem posed in \cite{karlsson2001montecarlo} consists in tracking two sources and estimating their unknown state vector, given their correct initial states $\vect{s}_0^1 = [6500, -1000, 2000, -50, 100, 0]$ and $\vect{s}_0^2 = [5050, -450, 2000, 100, 50, 0]$. We consider a more complex scenario by adding a third source, with initial state $\vect{s}_0^3 = [8000, 500, 2000, -100, 0, 0]$, which passes close to one of the other sources at a certain instant.

We apply the SIR/MCJPDA filter from \cite{karlsson2001montecarlo} and OMGP to perform data association on the observations. The SIR/MCJPDA filter consists of a set of joint particle filters that perform tracking of multiple sources, combined with a joint probability data association (JPDA) technique which provides instantaneous data association. The number of particles used in this experiment is $25000$. In order to operate correctly, the SIR/MCJPDA filter requires complete knowledge of the used state-space model and the initial state vectors $\vect{x}_0^i$. Note that OMGP is completely blind in this regard. The OMGP algorithm is operated first in its incremental online setting. For illustration purposes, we also include results of the batch version of the OMGP algorithm.

\begin{figure*}[ht]
\centerline{\subfloat[Trajectories identified by SIR/MCJPDA.]{\includegraphics[width=6.1cm]{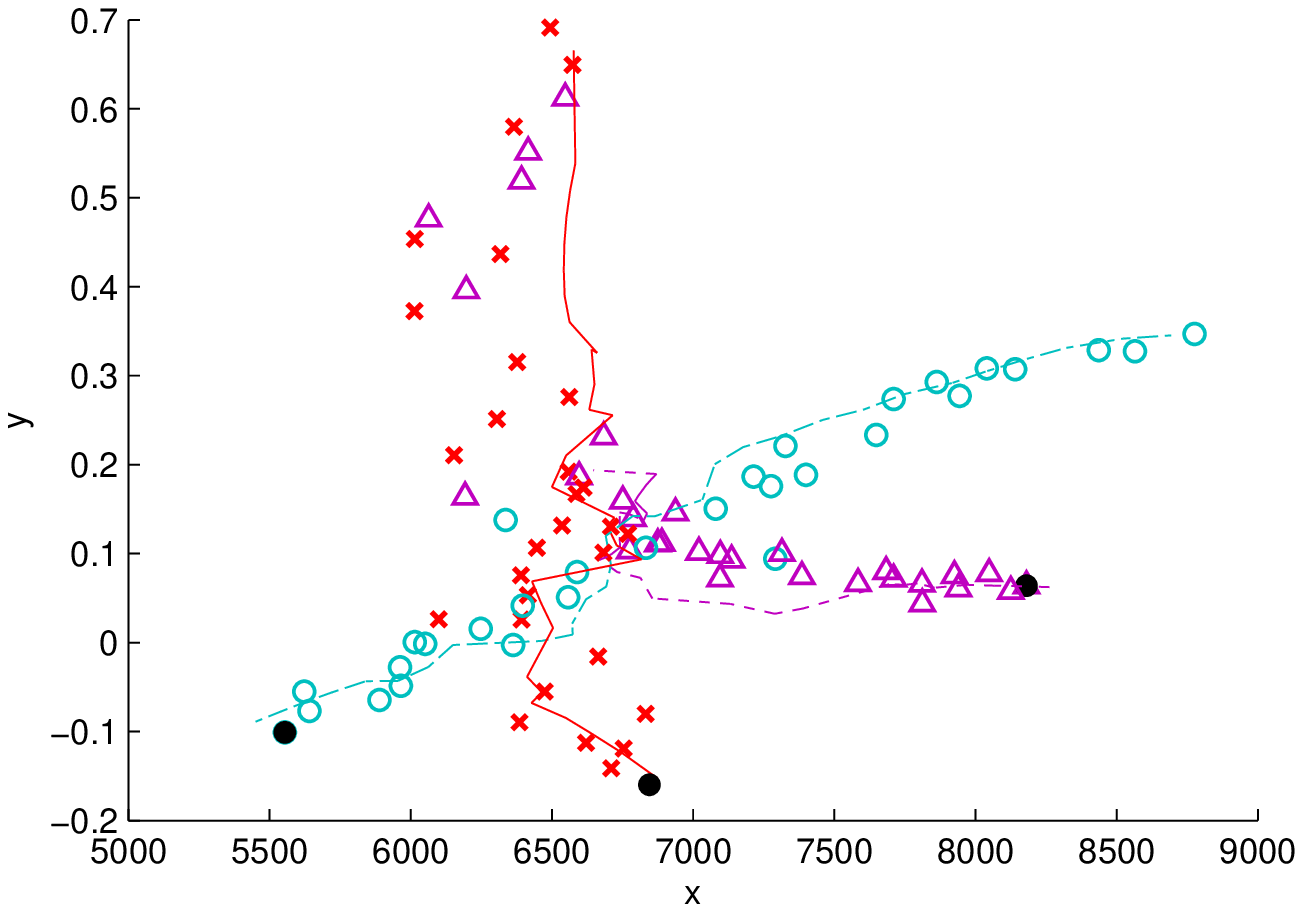} } \hfil
\subfloat[Trajectories identified by OMGP, incremental online version.]{\includegraphics[width=6.1cm]{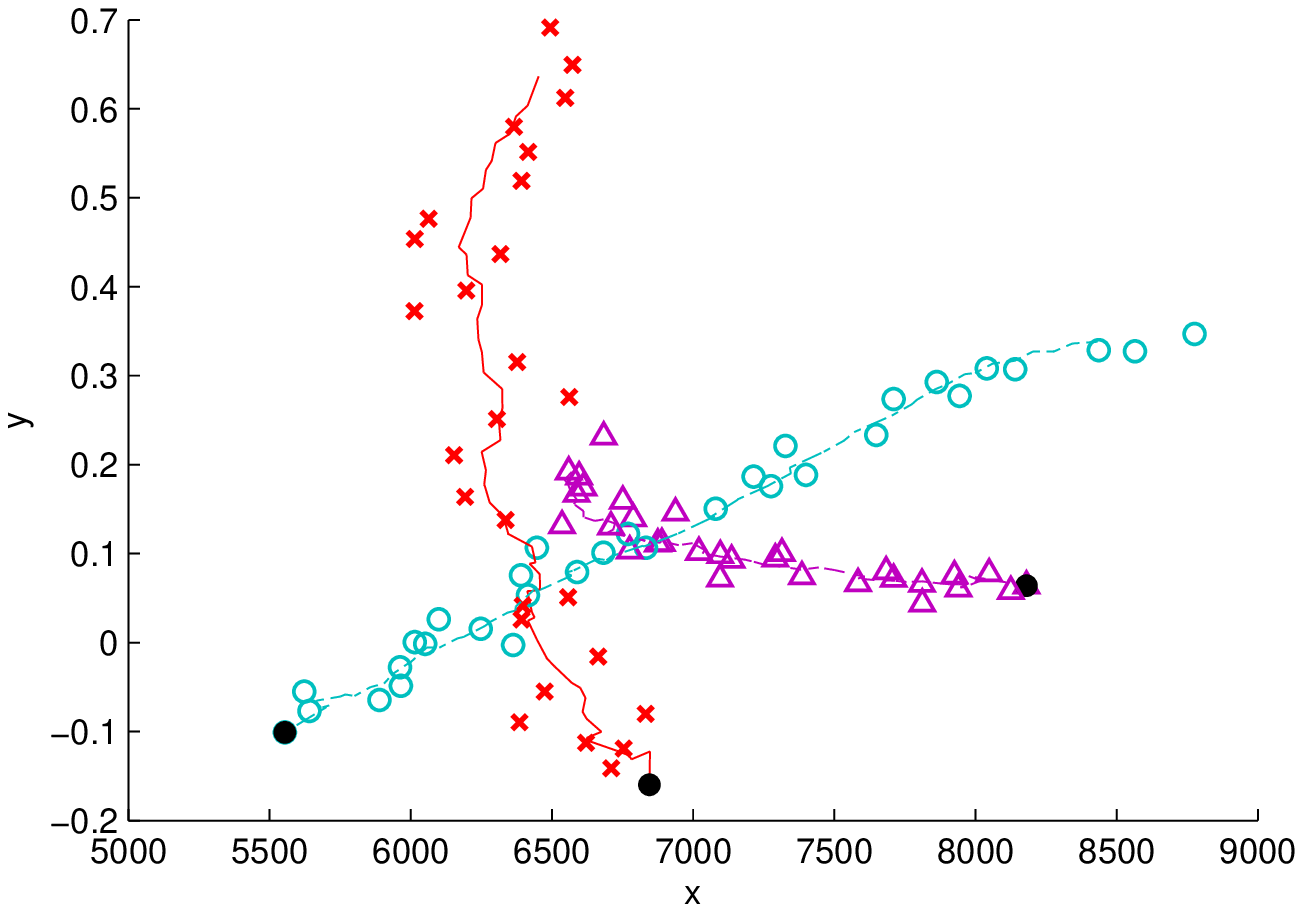} } }
\centerline{\subfloat[Trajectories identified by OMGP, batch solution.]{\includegraphics[width=6.1cm]{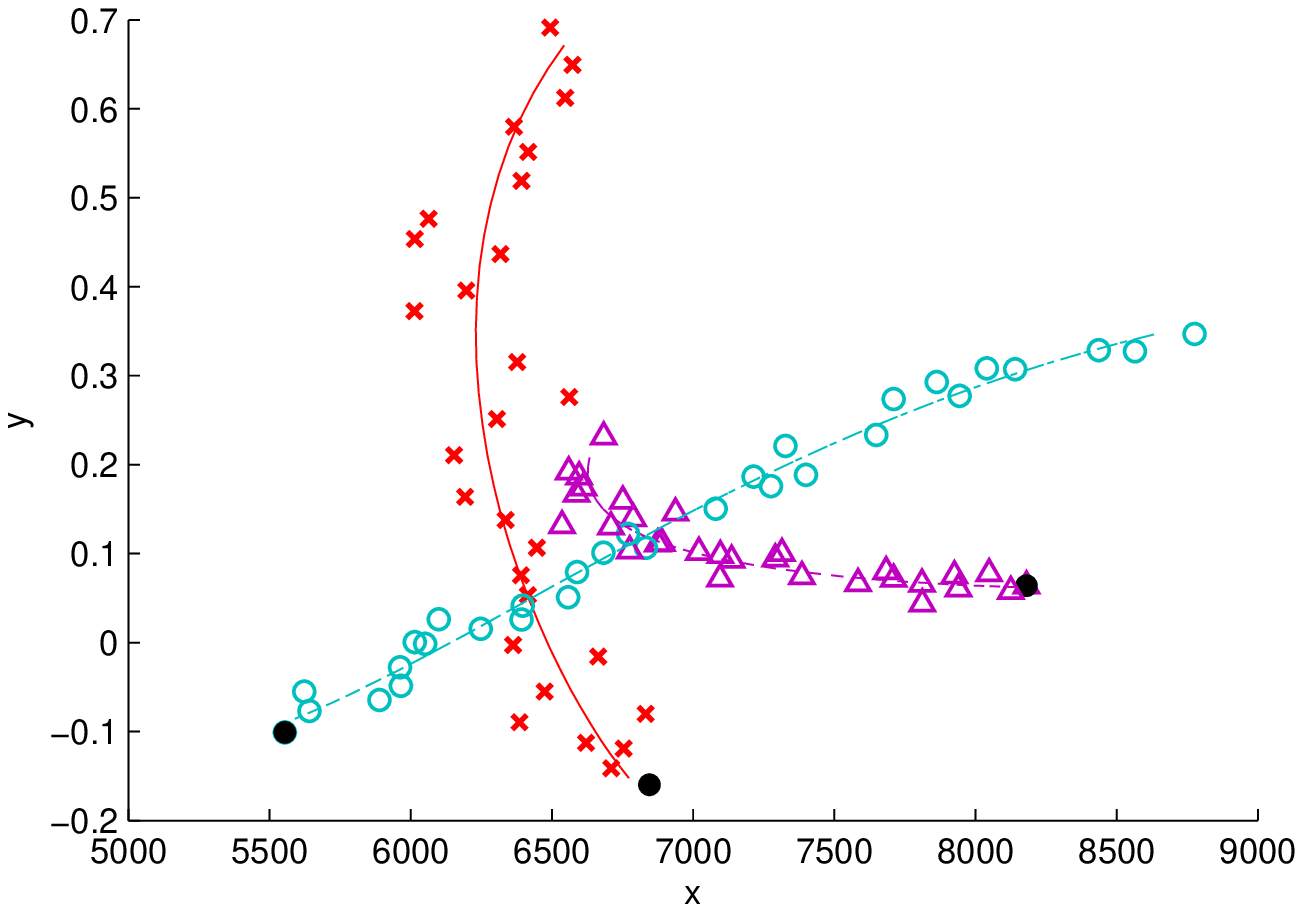} }}

\caption{Missile-to-air data association problem with three sources. The starting point of each source is marked with a black dot.}
\label{fig:mta}
\end{figure*}

The trajectories obtained by each method can be found in Fig.~\ref{fig:mta}, along with the predicted measurements. Although the SIR/MCJPDA filter initially performs correctly, it encounters difficulties at the point where the sources come close. After this point it shows erroneous assignments for at least one trajectory. Its mistakes are mainly due to its state vector depending only on $1$ previous state, which proves insufficient if the sources are close during multiple consecutive measurements. The online version of OMGP does not show this problem. The smoothest solution is obtained by batch OMGP, which performs a global evaluation of the entire trajectories.

To evaluate the performance of the algorithms, we measure the RMSE of each trajectory. These values can be found in Table~\ref{table:mta}, along with the number of observations that are assigned to the wrong trajectories, $n_{err}$, out of a total of $90$ observations. As can be observed, both versions of the OMGP algorithm obtain superior results compared to SIR/MCJPA. Furthermore, while SIR/MCJPDA requires complete knowledge of the state-space model and the initial state vectors, OMGP does not require any knowledge of the underlying model. %, due This is mainly due to its state vector depending only on the previous state. The assigns $17$ observations to different trajectories, compared to $1$ wrongly assigned observation for OMGP. The SIR/MCJPDA filter shows more errors when the observations are close, since it performs data association at each time instant and lacks knowledge of future data. The solution of OMGP, on the other hand, is based on a global evaluation of the entire trajectories. %Finally, notice that the SIR/MCJPDA filter has complete knowledge of the used state-space model and the initial state vectors $\vect{x}_0^i$ in this implementation, while OMGP is completely blind in this regard.
%: $err_{labels} = \frac{n_{err}}{n_{total}}$, where $n_{err}$ indicates erroneous assignments and $n$ the total number of observations.

\begin{table}
\caption{RMSE comparison on the missile-to-air data association problem.}
\begin{center}
\begin{tabular}{|l|l l l|c|}
  \hline
  % after \\: \hline or \cline{col1-col2} \cline{col3-col4} ...
  Algorithm & RMSE \#1 & RMSE \#2 & RMSE \#3 & $n_{err}$ \\
  \hline
  SIR/MCJPDA & 292.46 & 150.07 & 258.14 & 17 \\
  OMGP (online) & 182.31 & 151.46 & 163.92  & 6 \\
%  OMGP (online) & 217.62 & 137.42 & 131.06  & 3 \\
  OMGP (batch) & 133.30 & 80.23 & 118.94 & 1 \\
  \hline
\end{tabular}
\end{center}
\label{table:mta}
\end{table}

\subsubsection{Interference alignment in OFDM wireless networks}

Interestingly, the data association problem can be found in contexts that go beyond standard multi-target tracking scenarios, such as digital communications \cite{vanvaerenbergh2009path}. In the third experiment we apply OMGP to a data association problem that occurs in wireless communication networks.

Interference alignment (IA) is a concept that has recently emerged as a solution to raise the capacity of wireless multiple-input multiple-output (MIMO) networks %much more than what was previously believed to be possible
\cite{cadambe2008interference}. The underlying idea of IA along the spatial dimensions is that the interference from other transmitters must be aligned at each receiver in a subspace orthogonal to the signal space. %in a lower-dimensional subspace. %Once all the interfering signals lie in the same subspace, interference can be removed through zero-forcing filtering. %The aligning itself is performed by multiplying the outputs of each transmit antenna with a properly chosen complex scalar.
%Several algorithms have been developed in order to obtain valid IA solutions for a particular scenario.
%When multiple subcarriers are used, a digital filter must be designed
In order to implement interference alignment in scenarios with multiple subcarriers, a digital filter must be applied at each transmit antenna. Here we will consider a 3-user interference channel with two antennas per node and OFDM modulation using $N_c$ subcarriers \cite{proakis1995digital}, which allows for two possible filter responses per subcarrier. %(i.e. a total of $2^{N_c}$ potential solutions).
%At each subcarrier frequency, the frequency response of this filter must correspond to one of the $2$ possible values.
Since only smooth frequency responses can be implemented, the smoothest solution of the $2^{N_c}$ possible choices should be selected.

%These solutions vary smoothly over neighboring subcarriers due to small channel differences.

%In case a communication link with multiple subcarriers is used, for instance by applying OFDM modulation \cite{proakis1995digital},

%However, only two of the $2^{N_c}$ possible choices are smooth and, in consequence, interesting in practice.

\begin{figure*}[h!]
\centerline{\subfloat[]{\includegraphics[width=6.1cm]{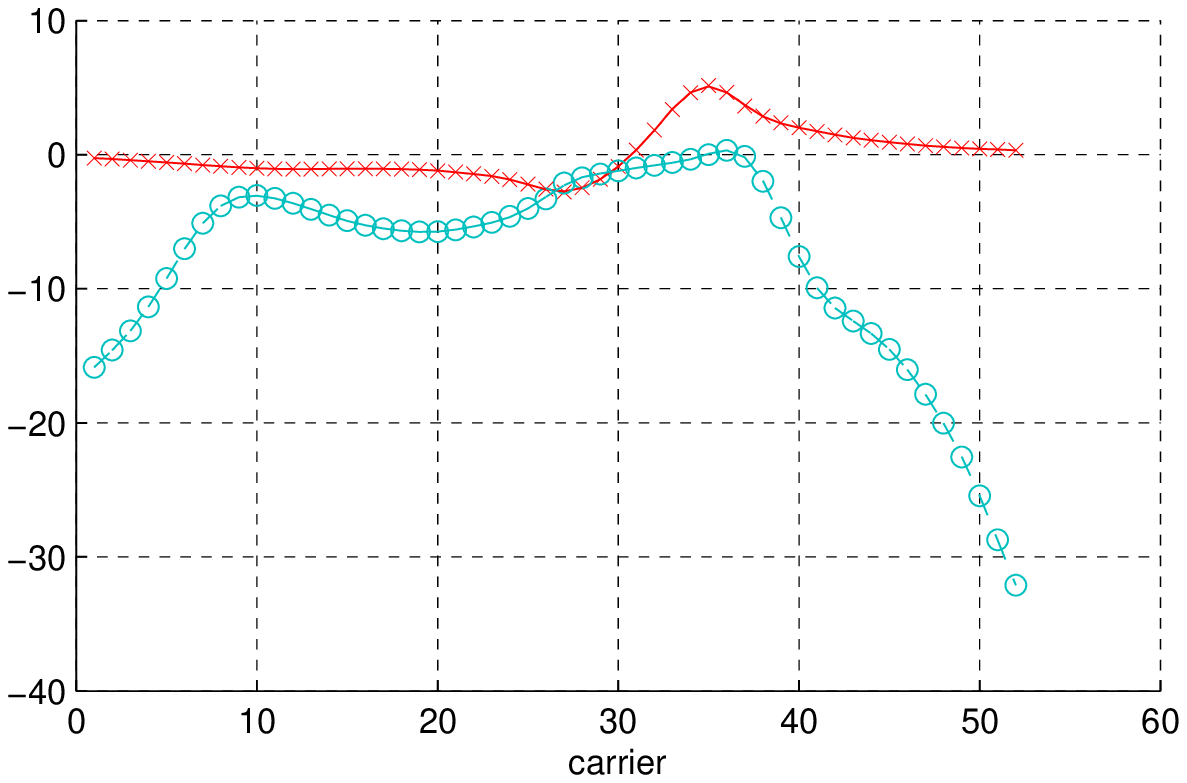} } \hfil
\subfloat[]{\includegraphics[width=6.1cm]{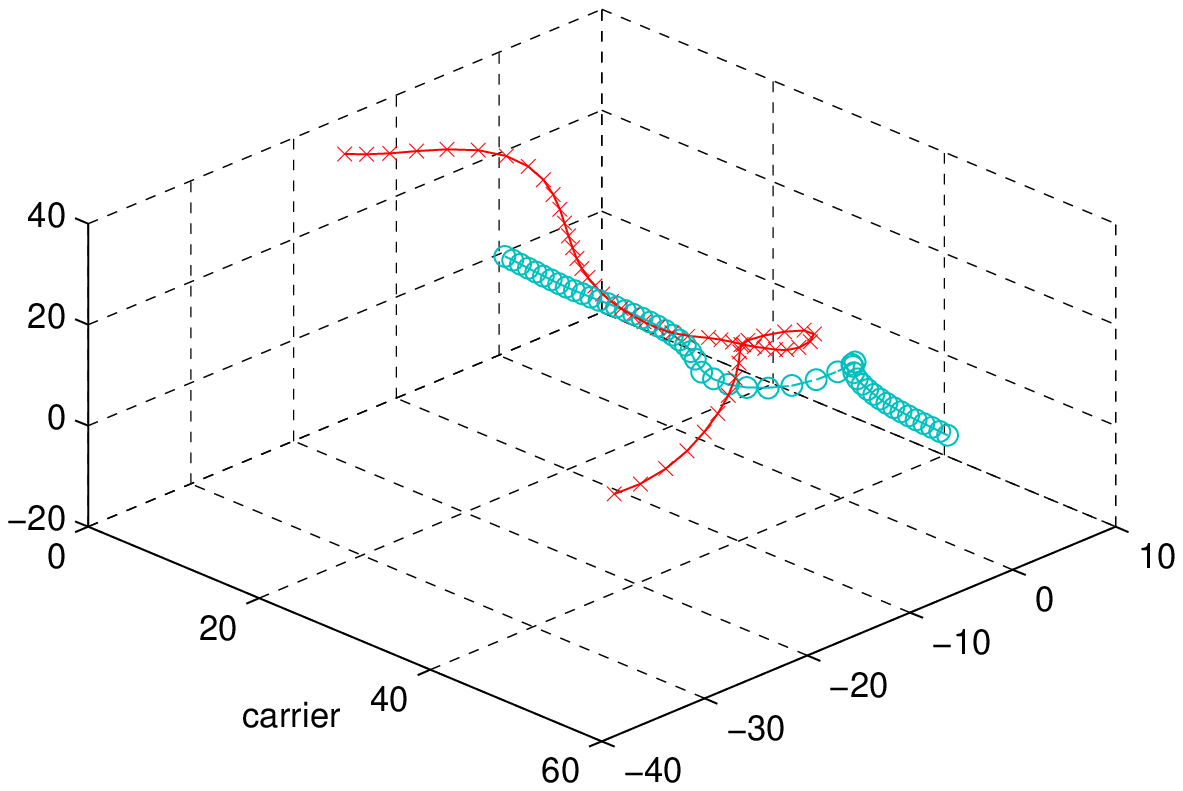} } }
\centerline{\subfloat[]{\includegraphics[width=6.1cm]{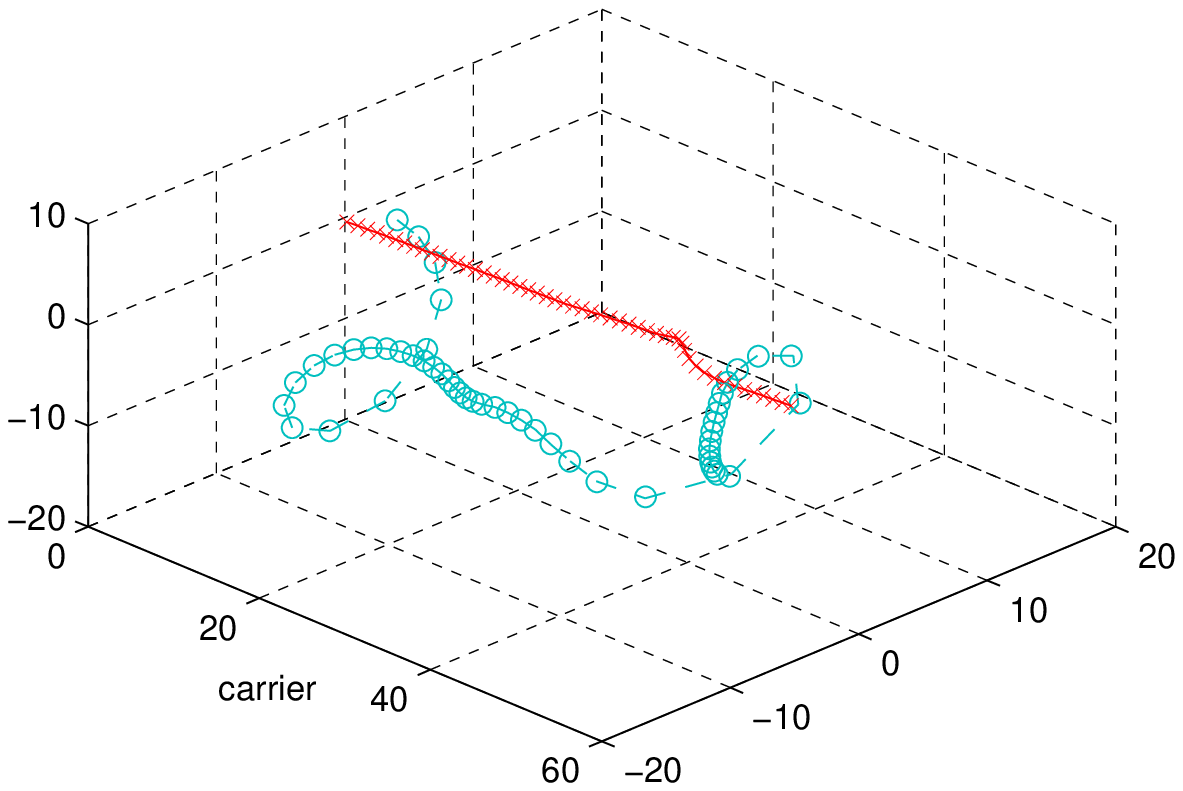} } \hfil
\subfloat[]{\includegraphics[width=6.1cm]{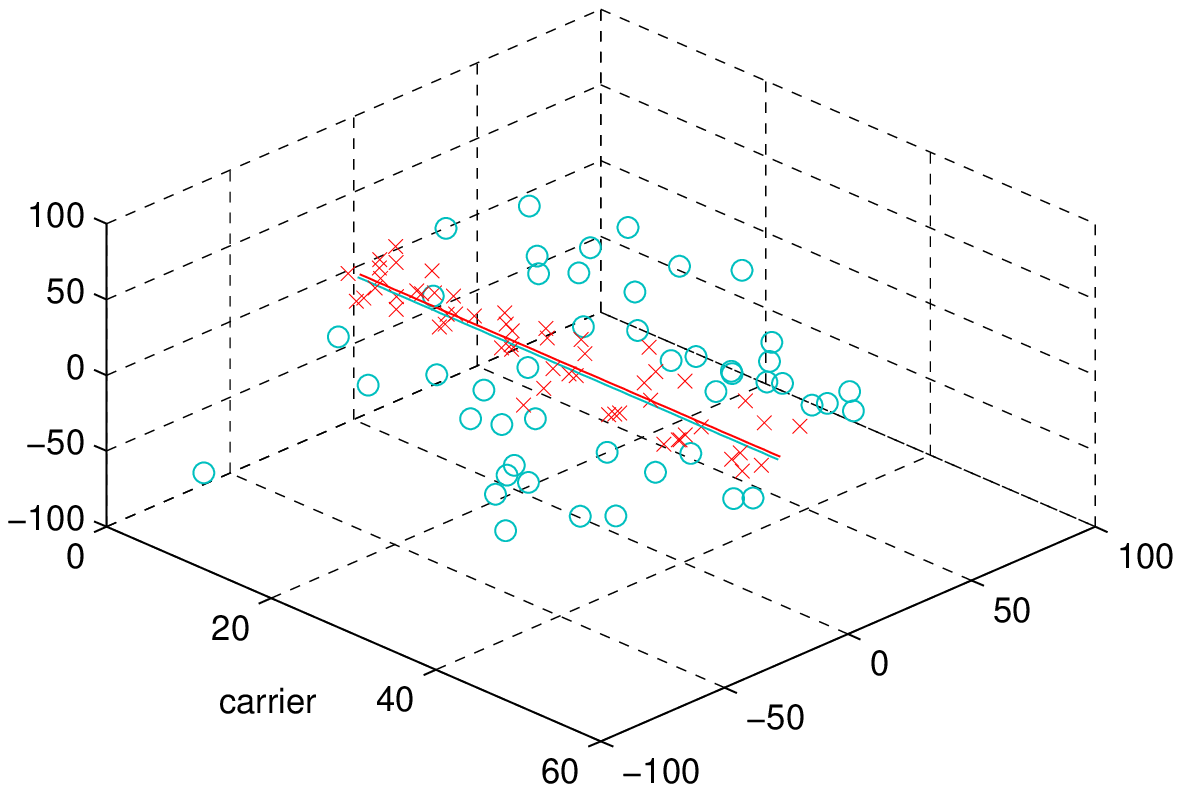} } }
\caption{Data association results obtained by OMGP on different interference alignment problems. (a) shows the IA solutions (imaginary part only) for the first simulated data set. (b) and (c) show the solutions for the first and second simulated data sets (real and imaginary part versus subcarrier number). (d) shows the IA solution for a real-world data set. Note that complex values are simply simply treated as two-dimensional real data in this experiment.}
\label{fig:interference}
\end{figure*}

This combinatorial problem corresponds to a data association problem in which only the smoothest curve is of interest. %data is taken from two process of which at least one can be represented by a smooth curve
(see Fig.~\ref{fig:interference}(a)). The data used for this experiment consists of two simulated data sets and one data set obtained with a MIMO test bed setup\footnote{See \cite{gutierrez2011frequency} for a full description of the used test bed.}, each using $52$ subcarriers. In Fig.~\ref{fig:interference} we illustrate the solutions obtained by OMGP on these data sets. While the simulated data sets from Fig.~\ref{fig:interference}(b) and Fig.~\ref{fig:interference}(c) represent reasonably simple data association problems, the performance of OMGP on the real-world data set of Fig.~\ref{fig:interference}(d) shows that it is capable of correctly distinguishing the smoothly-varying solution from the surrounding noisy data. As a matter of fact, we have been able to successfully implement OMGP in the IA setting for a parallel ungoing research project.

\subsection{Regression tasks}

We now consider application of the model in more standard regression tasks. In particular, we consider tasks where the target density is multimodal, which is the case when the data comes from multiple sources.

\begin{figure*}[!htb]
\centerline{\subfloat[Original data set.]{\includegraphics[width=6.1cm]{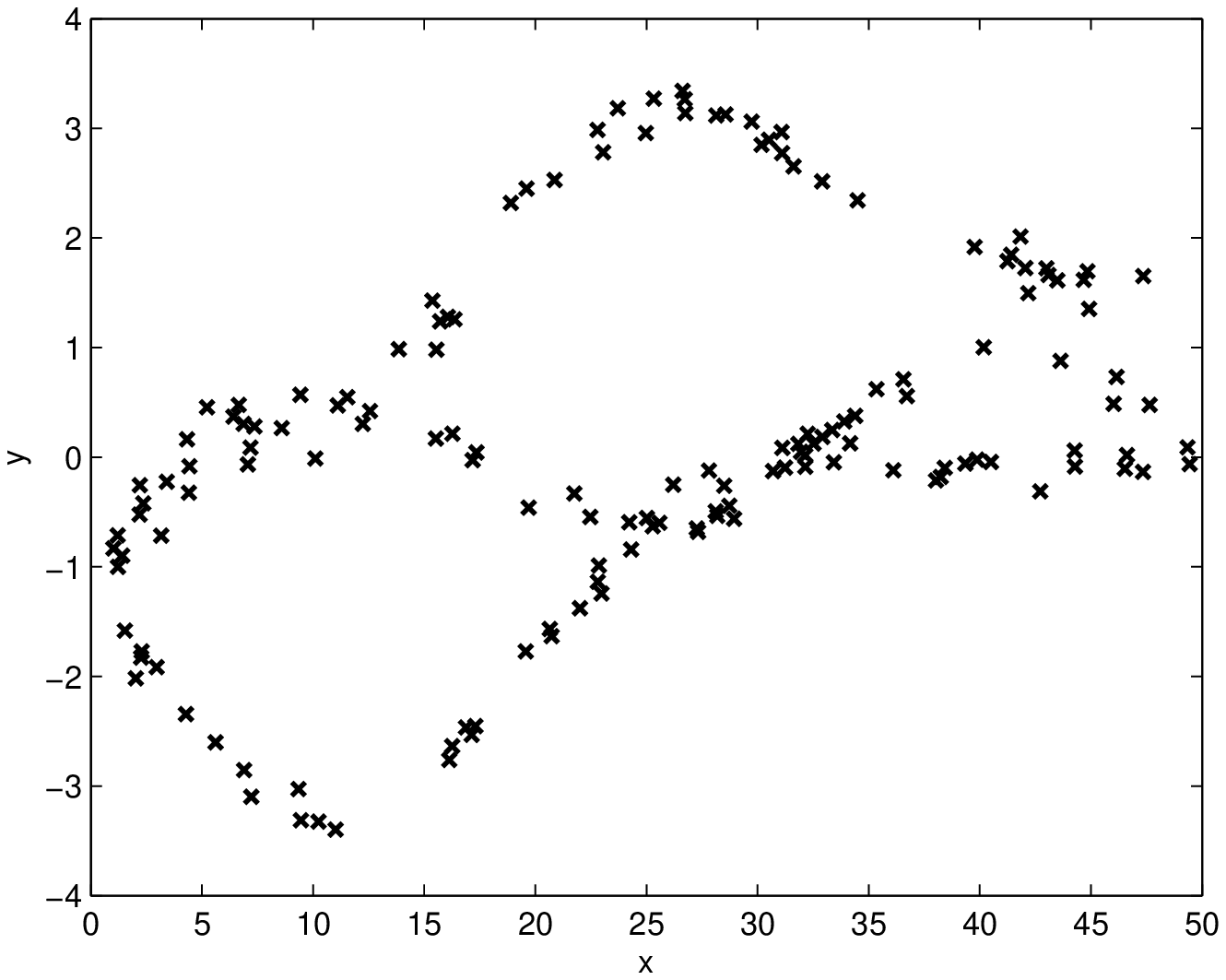}
}
\hfil
\subfloat[Inferred labels and  predictive log-probs.]{\includegraphics[width=6.1cm]{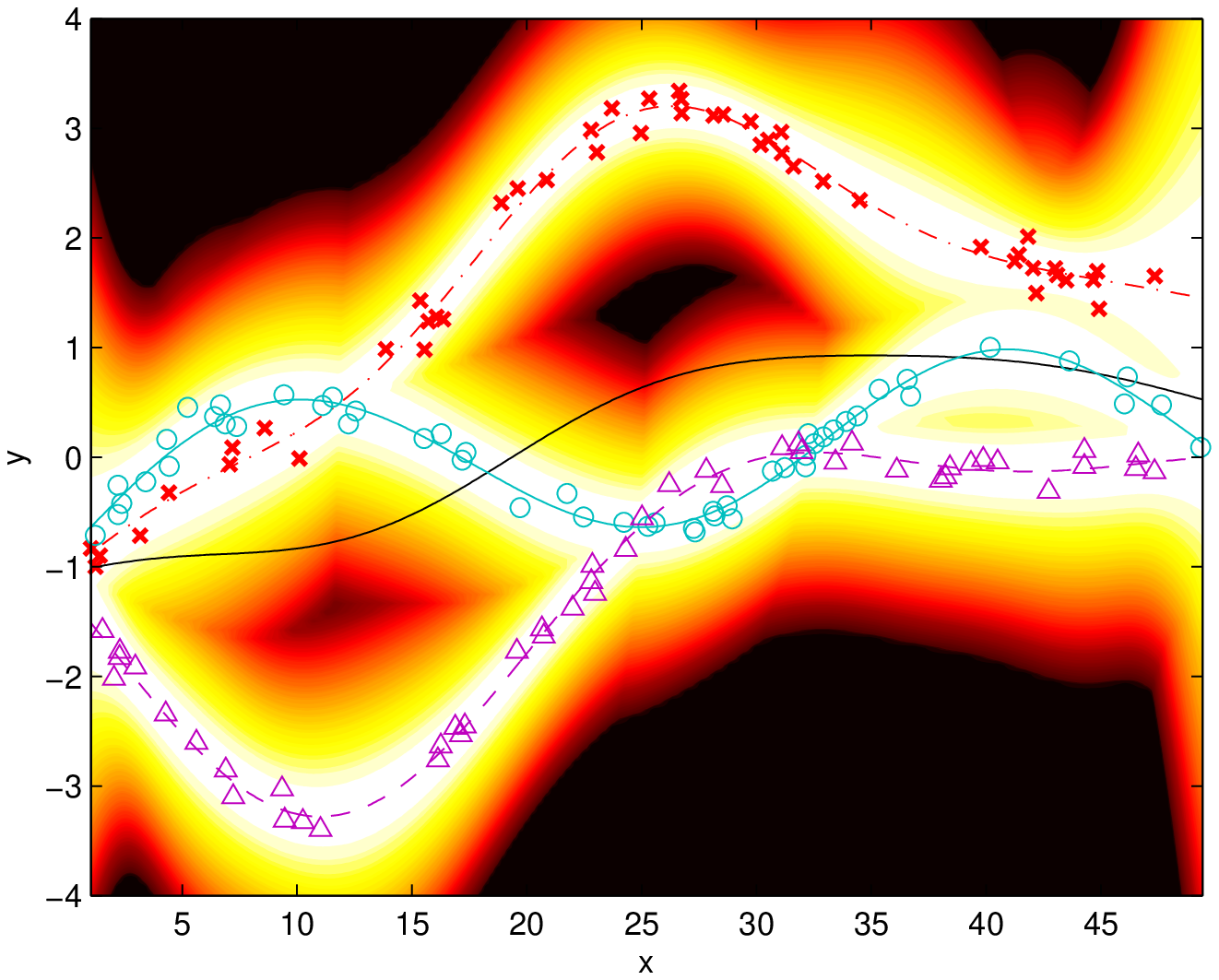}
}}
\caption{Posterior log-probability of the OMGP model and label inference.}
\label{fig:toy2d}
\end{figure*}
\subsubsection{Multilevel regression} Consider the data set from Fig.~\ref{fig:toy2d}(a), which corresponds to observations from three independent functions. A normal GP would fail to produce valid multimodal outputs and previously proposed mixtures of GPs would restrict the component GPs to local parts of the space. OMGP can properly label each observation according to the generating function and provide multimodal predictive distributions, as depicted in Fig.~\ref{fig:toy2d}(b).

Fig.~\ref{fig:toy2d} can also be interpreted as measurements of the position of three particles moving along one dimension, of which snapshots are taken at irregular time intervals (horizontal axis). Each snapshot introduces noise in the position measurement and does not necessarily capture the position of all the particles. In this case OMGP could be used to predict the position of any particle at any given point in time, as well as to properly label the samples in each snapshot.

\begin{figure*}[!htb]
\centerline{\subfloat[Noisy sinc. Standard GP]{\includegraphics[width=6.1cm]{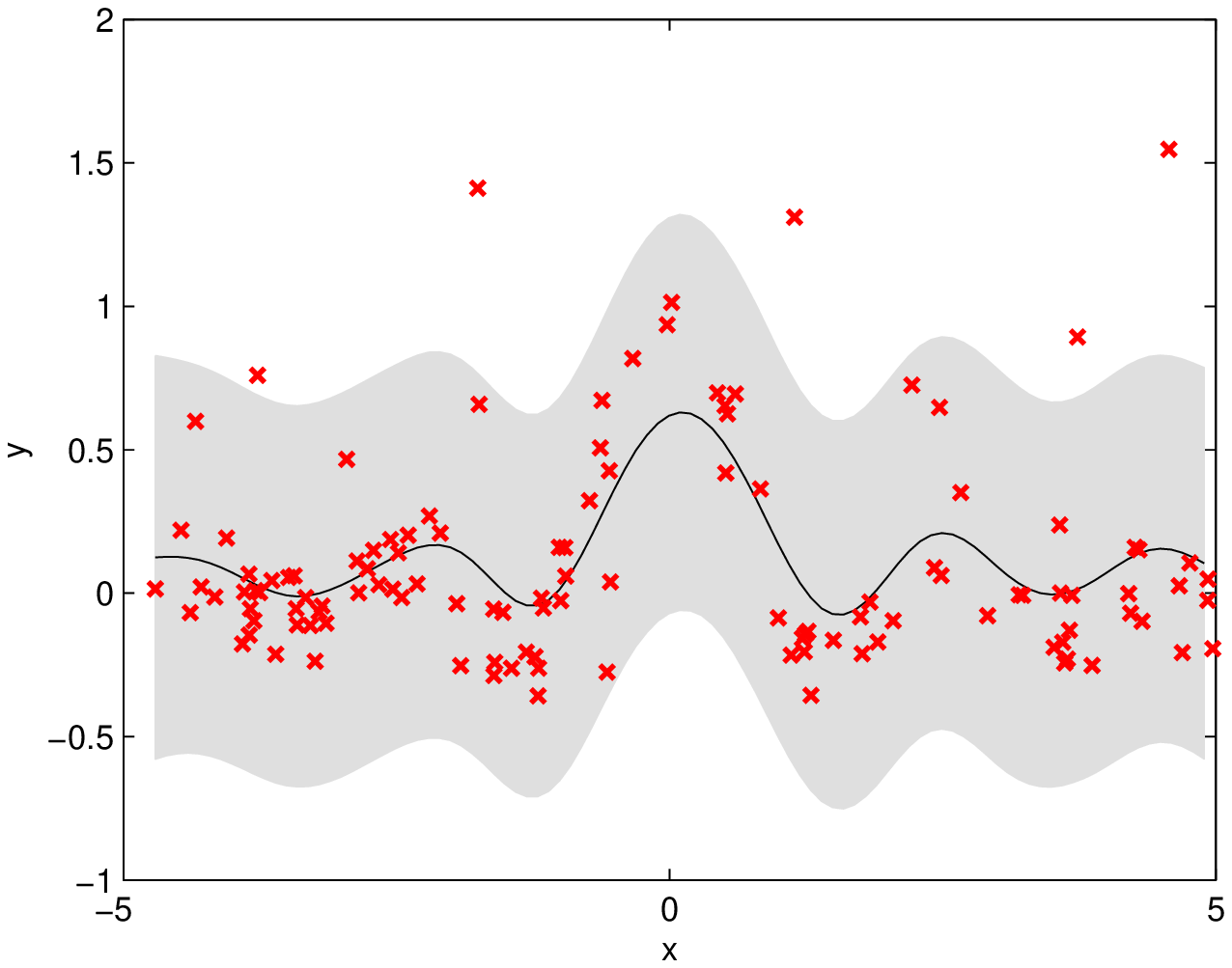}
}
\hfil
\subfloat[Noisy sinc. OMGP]{\includegraphics[width=6.1cm]{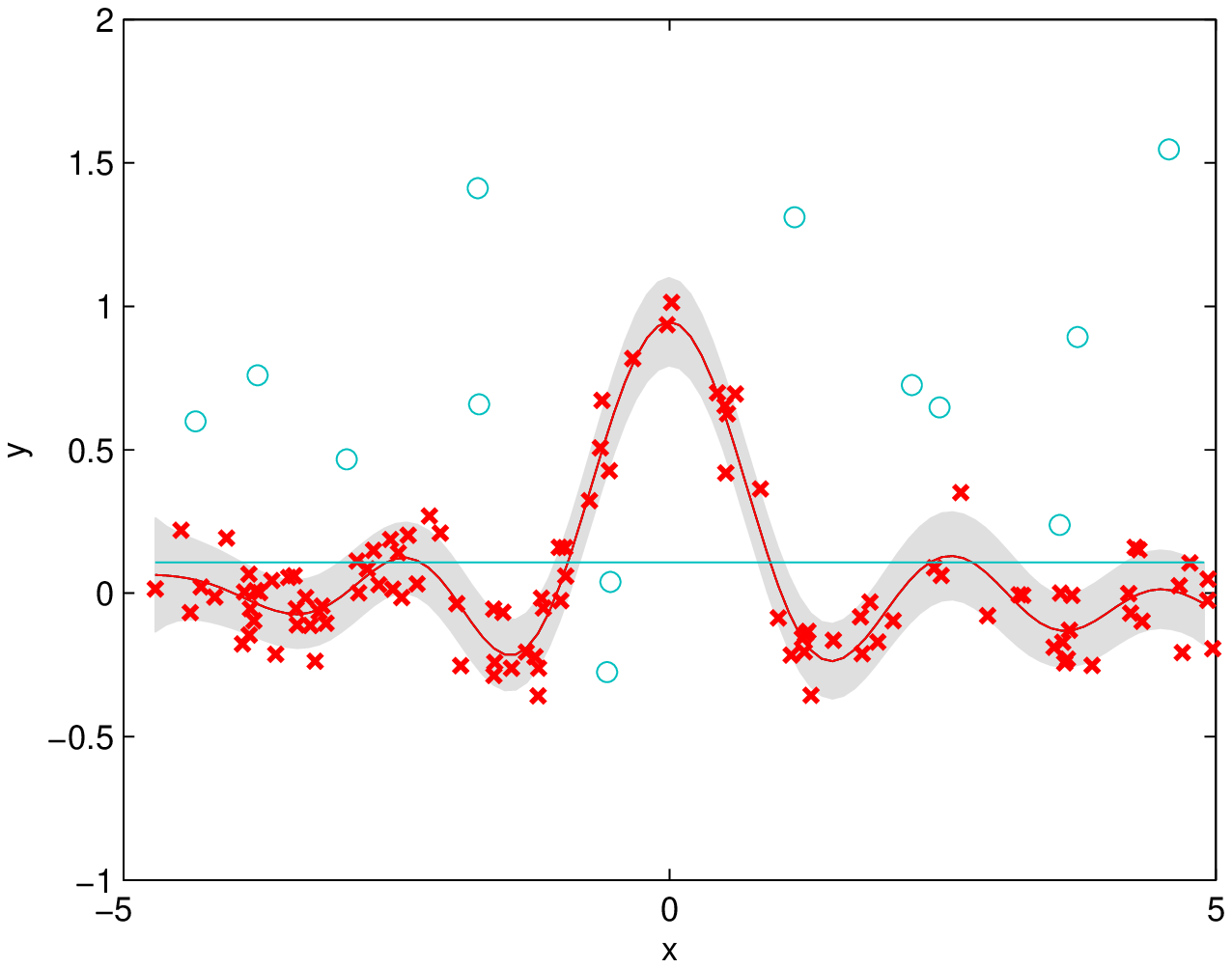}
}}
\centerline{\subfloat[Motorcycle. Standard GP]{\includegraphics[width=6.1cm]{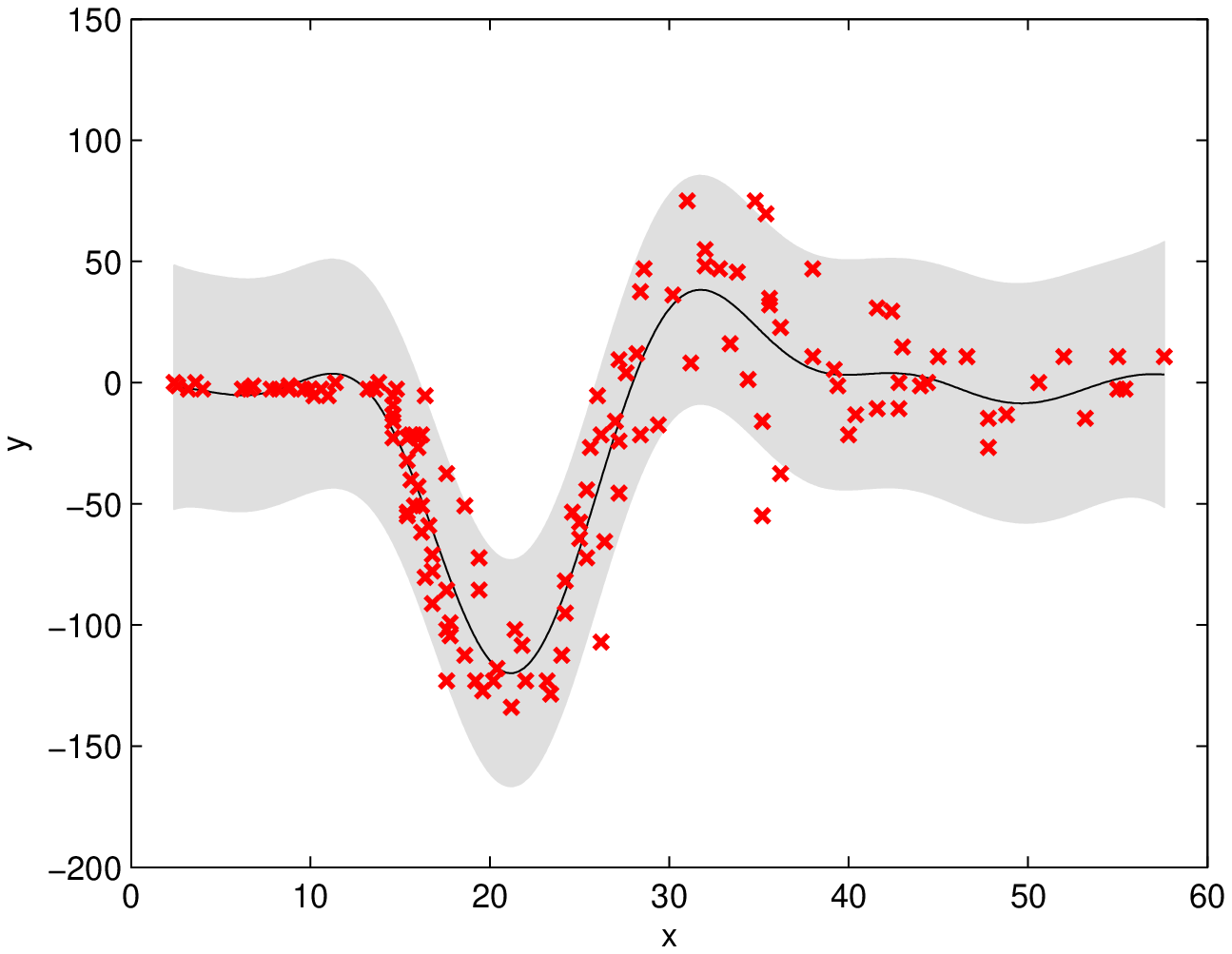}
}
\hfil
\subfloat[Motorcycle. OMGP]{\includegraphics[width=6.1cm]{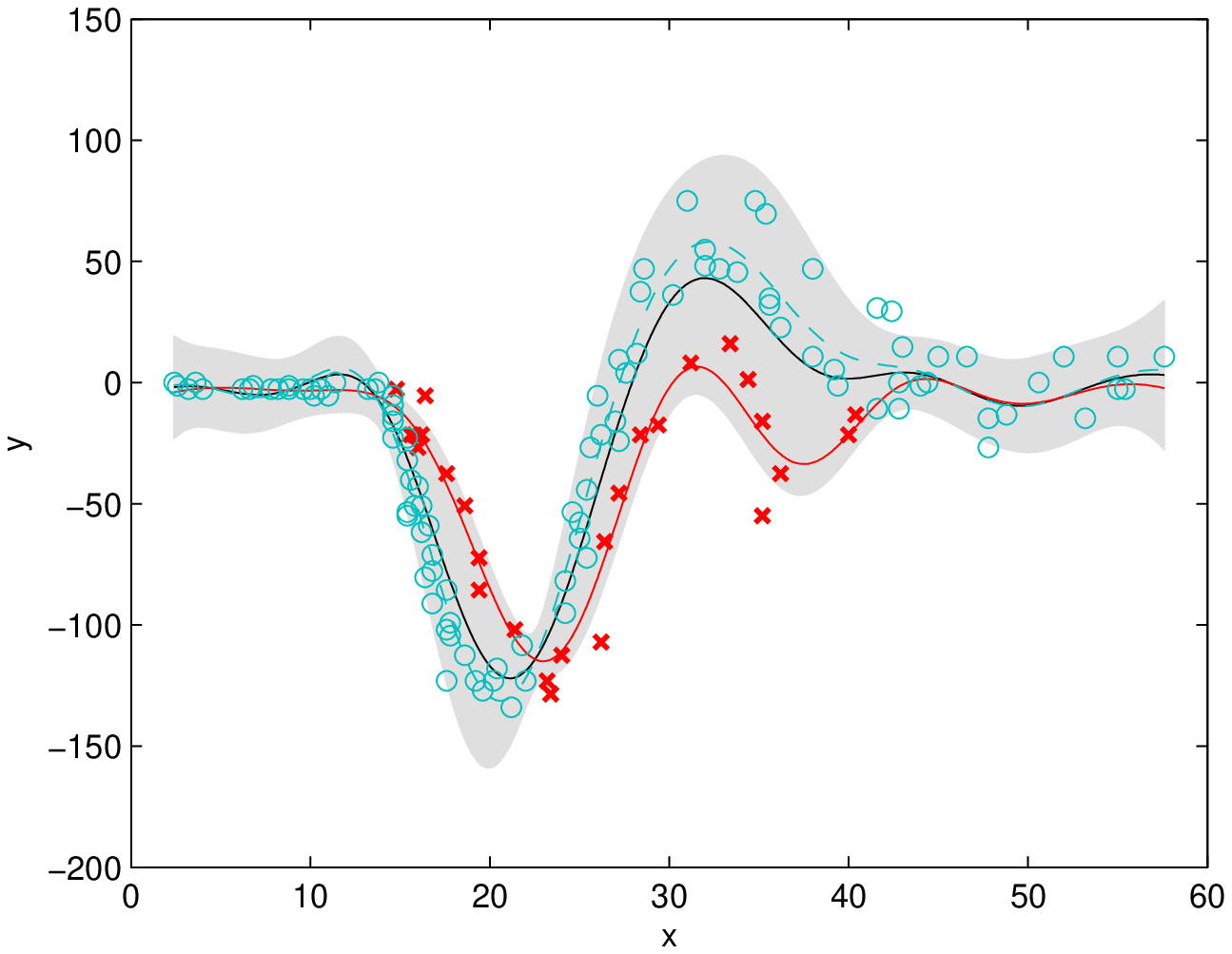}
}}
\caption{Predictive means and variances for two different data sets. The shaded area denotes $\pm2$ standard deviations around the mean. Top row: Noisy sinc with outliers. (a) Standard GP and (b) OMGP with a noise-only component. (Only the predictive mean and variance of the signal component is depicted, which includes noise $\sigma^2$). Bottom row:  Silverman's motorcycle data set. }
\label{fig:sincmoto}
\vspace{-0.2cm}
\end{figure*}
\subsubsection{Robust regression} Since each GP in the mixture can use a different covariance function, it is possible to use a GP to capture unrelated outliers and another one to interpolate the main function. This is easily achieved by a mixture of two GPs, one with the ARD-SE covariance function and another with $k(x,x')=b^2\delta(x,x')$, i.e., white noise. We consider the problem of regression in a noisy sinc in which some outliers have been introduced in Fig.~\ref{fig:sincmoto} (top row). Observe how OMGP both identifies the outliers and ignores them, resulting in much better predictive means and variances.

%\begin{figure*}[!htb]
%\centerline{\subfloat[Standard GP]{\includegraphics[width=6.1cm]{motogp}
%}
%\hfil
%\subfloat[OMGP]{\includegraphics[width=6.1cm]{moto}
%}}
%\caption{Predictive means and variances for Silverman's motorcycle data set. The shaded area denotes $\pm2$ standard deviations around the mean.}
%\label{fig:moto}
%\end{figure*}
%

\subsubsection{Heteroscedastic behavior} Finally, Fig.~\ref{fig:sincmoto} (bottom row) shows the results of running a GP and OMGP on the motorcycle data set from \cite{silverman}. Two components have been identified, which might or might not correspond to two actual physical mechanisms alternatively producing observations. The predictive variances show improved behavior with respect to the standard GP.

\section{Discussion and future work}
\label{sec:conclusions}

In this work we have introduced a novel GP mixture model inspired by multi-target tracking problems. The new model has the important difference with respect to previous approaches of using global mixture components and assigning samples to components by relying on their value in output space, instead of input space (as it is done when gating functions are used).

A simple and efficient algorithm for inference relying on the variational Bayesian framework has been provided. The model can be applied in practice due to the use of an improved, KL-corrected variational bound to learn the hyperparameters. Direct optimization of this bound both to obtain an approximate posterior and to learn the hyperparameters will be considered in a further work.

The OMGP model offers promising results when tracking moving targets, as has been illustrated experimentally in Section \ref{sec:experiments} and compares favorably with established methods in the field. Also, through imaginative application of the model using different covariance functions we were able to adapt the approach to robust regression and heteroscedastic noise.

Naive implementation of GPs limits their applicability to only a few thousand data samples. However, recent advances in sparse approximations (e.g. \cite{snelson06sparse,titsias}) greatly should enable our approach to be applied to much larger data sets.

\section{Acknowledgments}
The authors wish to thank Oscar Gonz\'alez, University of Cantabria, for providing the data used in the interference alignment experiment. The first and second authors were supported by MICINN (Spanish Ministry for Science and Innovation) under grants TEC2010-19545-C04-03 (COSIMA) and CONSOLIDER-INGENIO 2010 CSD2008-00010 (COMONSENS). Additionally, funding to support part of this collaborative effort was provided by PASCAL's Internal Visiting Programme.

\bibliographystyle{model1-num-names}
\bibliography{omgp}

%% Authors are advised to submit their bibtex database files. They are
%% requested to list a bibtex style file in the manuscript if they do
%% not want to use model1a-num-names.bst.

%% References without bibTeX database:

% \begin{thebibliography}{00}

%% \bibitem must have the following form:
%%   \bibitem{key}...
%%

% \bibitem{}

% \end{thebibliography}

\end{document}